\def\debug{0}

\documentclass{article}

\PassOptionsToPackage{numbers, compress}{natbib}

\usepackage[preprint]{neurips_2024}

\usepackage[T1]{fontenc}
\usepackage[utf8]{inputenc}
\usepackage{verbatim}
\usepackage{refstyle}
\usepackage{mathtools}
\usepackage{amsmath}
\usepackage{amssymb}
\usepackage{graphicx}

\usepackage{float} 

\usepackage{titletoc} 

\usepackage{hyperref}       
\usepackage{url}            
\usepackage{booktabs}       
\usepackage{amsfonts}       
\usepackage{nicefrac}       
\usepackage{microtype}      
\usepackage{xcolor}         

\usepackage{multirow}

\usepackage{amsmath,amsfonts,bm}

\def\Figref#1{Fig.~\ref{#1}}

\def\eqref#1{equation~\ref{#1}}
\def\Eqref#1{Eq.~\ref{#1}}

\def\1{\bm{1}}

\DeclareMathAlphabet{\mathsfit}{\encodingdefault}{\sfdefault}{m}{sl}
\SetMathAlphabet{\mathsfit}{bold}{\encodingdefault}{\sfdefault}{bx}{n}

\title{Dissecting the Interplay of Attention Paths in a Statistical Mechanics Theory of Transformers}

\author{%
Lorenzo Tiberi$^{1, 2}$ \quad Francesca Mignacco$^{3,4}$ \quad Kazuki Irie$^{1, 2}$ \quad Haim Sompolinsky$^{1,2, 5}$ \\  
$^1$Center for Brain Science, Harvard University, Cambridge, MA, USA\\
$^2$Kempner Institute for the Study of Natural and Artificial Intelligence,\\
Harvard University, Cambridge, MA, USA\\
$^3$Graduate Center, City University of New York, NY, USA\\
$^4$Joseph Henry Laboratories of Physics, Princeton University, NJ, USA\\
$^5$Edmond and Lily Safra Center for Brain Sciences,\\
Hebrew University of Jerusalem, Jerusalem, Israel\\
  \texttt{ltiberi@fas.harvard.edu}, \texttt{fmignacco@princeton.edu} \\
  \texttt{kirie@fas.harvard.edu}, \texttt{hsompolinsky@mcb.harvard.edu} \\
}

\newcommand{\Fnotes}[1]{\ifnum\debug=1{\color{teal} [FM: #1]}\fi}
\newcommand{\Lnotes}[1]{\ifnum\debug=1{\color{violet} [LT: #1]}\fi}
\newcommand{\Knotes}[1]{\ifnum\debug=1{\color{blue} [KI: #1]}\fi}

\makeatletter

\begin{document}

\maketitle

\begin{abstract}
Despite the remarkable empirical performance of transformers, their
theoretical understanding remains elusive. Here, we consider a deep
multi-head self-attention network, that is closely related to transformers
yet analytically tractable. We develop a statistical mechanics theory
of Bayesian learning in this model, deriving exact equations for the
network's predictor statistics under the finite-width thermodynamic
limit, i.e., $N,P\rightarrow\infty$, $P/N=\mathcal{O}(1)$, where
$N$ is the network width and $P$ is the number of training examples.
Our theory shows that the predictor statistics are expressed as a
sum of independent kernels, each one pairing different \textit{attention
paths}, defined as information pathways through different attention
heads across layers. The kernels are weighted according to a \textit{task-relevant
kernel combination} mechanism that aligns the total kernel with
the task labels. As a consequence, this interplay between attention
paths enhances generalization performance. Experiments confirm our
findings on both synthetic and real-world sequence classification
tasks. Finally, our theory explicitly relates the kernel combination
mechanism to properties of the learned weights, allowing for a qualitative
transfer of its insights to models trained via gradient descent. As
an illustration, we demonstrate an efficient size reduction of the network, by pruning those attention heads 
that are deemed less relevant by our theory.\footnote{Our code is public: \href{https://github.com/tiberilor/attention-paths-interplay}{https://github.com/tiberilor/attention-paths-interplay}}
\end{abstract}

\section{Introduction}
In recent years, transformer models  based on multi-head self-attention
layers \citep{vaswani2017attention,kim2016structured,ParikhT0U16,cheng16,lin2017structured,bahdanau2014neural} have achieved remarkable performance at natural language processing
and vision tasks \citep{gpt3,dosovitskiy2020image,devlin2019bert}. Yet, theoretical characterizations accounting for
the success of these architectures remain sparse.
Two fundamental
questions remain to a large extent unsolved: First, interpretability---how
can we discern task-relevant structures within the learned weights?
Second, generalization---what specific aspects of the transformer
architecture are responsible for their effective learning?
We posit that one important feature of transformers is the combination
of layer-wise multi-head organization with depth.
This provides the network
with a large number of \textit{attention paths}, defined as specific sequences
of heads through the attention layers.
Their \textit{interplay} is still
poorly understood by deep learning theory.

In most cases, theoretical characterizations of transformers' expressivity \cite{fu2024can}, inductive bias \cite{sahiner2022unraveling,tarzanagh2023max}, generalization \cite{cui2024phase, rende24, noci2024shaped} and training dynamics \cite{li2022theoretical,boix2023transformers,tian2023scan} rely on simplifying assumptions on the network architecture. A characterization of attention paths is inaccessible in these models,
either because attention paths are not defined in the first place, as in models consisting of a single-head \cite{geshkovski2023mathematical, noci2024shaped}, a single-layer \cite{fu2024can,sahiner2022unraveling}, or both \cite{tarzanagh2023max,cui2024phase,rende24,li2022theoretical,tian2023scan, li2024}, or because the interplay between paths cannot be fully described due to constraints imposed on the learnable weights \cite{boix2023transformers}. A few works consider a multi-head, multi-layer architecture \cite{hahn2020theoretical, edelman2022inductive, nichani2024, reddy2024}, but address different questions than the present study, such as expressivity, generalization bounds, or phenomenological models. Further details on these and analogous works are discussed in Appendix~\ref{part:related_works}.

One characterization of the complete transformer architecture has been obtained in the Bayesian framework under the \emph{infinite-width} thermodynamic limit $N\to\infty$ (and infinite number of heads  $H\to\infty$) \citep{hron2020infinite,lavie2024towards}, an actively studied regime in which neural networks become equivalent to Gaussian processes (GP) \citep{lee2018deep,matthews2018gaussian}.
However, the attention
paths interplay is lost in this limit because the network's hidden weights
remain statistically independent after learning. This limitation can be overcome by considering
the \emph{finite-width} thermodynamic limit \citep{li2021statistical,hanin2023bayesian,pacelli2023statistical,pmlr-v202-cui23b}, where also the number of examples
$P\to\infty$ such that $P/N\to\alpha\in\mathbb{R}^{+}$. In this
regime, for example, multi-gated deep networks showcase task-relevant
interplay between gates, mediated by the learned weights \citep{li2022globally}. 

In this work, we apply the statistical mechanics theory of finite-width networks
to a deep multi-head self-attention model, which closely mimics the
attention paths interplay in transformers, while remaining analytically
tractable. Our main contributions can be summarized as follows:
\begin{itemize}
\item We derive exact equations for the predictor statistics under Bayesian
learning of the network's value weights, at fixed query
and key weights.
\item We shed light on the interplay between attention paths by uncovering
a \emph{task-relevant kernel combination} mechanism, emerging beyond
the GP limit ($\alpha>0$). This constructs the network's mean predictor as
an optimally weighted sum of many ``path-path kernels'', defined as similarity matrices between pairs of attention
paths, thereby improving generalization. 
\item We provide interpretability to this mechanism, by directly relating
it to the magnitude and correlations developed by the learned weights.
This allows our insights to be transferred outside the Bayesian framework, to networks trained with gradient descent.
As an application, we show that a trained network
can be reduced in size with minimal performance loss, by pruning those attention paths
that are deemed less relevant by our theory.
\item We corroborate our findings on both synthetic and real-world sequence
classification tasks, illustrating the two main benefits of kernel
combination: task-relevant weighting and correlation of the attention
paths, respectively.
\end{itemize}

\section{Model\label{sec:Model}}

\begin{figure}
\begin{centering}
\includegraphics[width=0.95\textwidth]{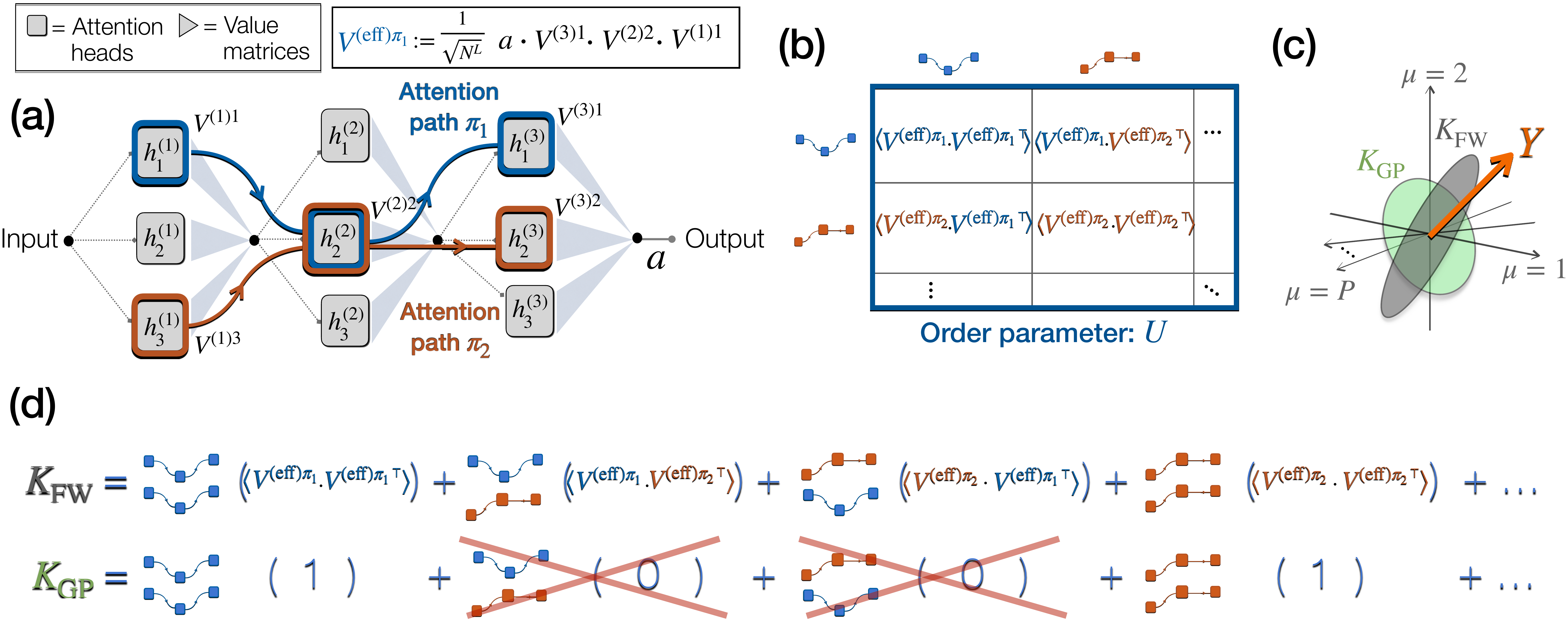}
\par\end{centering}
\caption{\label{fig:scheme}{\bf Scheme of the model and theory} {\bf (a)} Scheme of the model in terms of attention paths.
{\bf (b)} The order parameter assigns to each pair of paths a weight, given by the overlap between the corresponding effective weights. {\bf (c)} Alignment of the kernel PCs with the vector of task labels $Y$, in the finite-width (FW) vs GP regimes. {\bf (d)} Kernel as the weighted sum of many path-path kernels. Task-relevant kernel combination occurs in the finite-width regime (FW), but not in the GP limit, in which cross-path kernels are discarded, and same-path kernels are equally weighted. The result is an improved kernel-task alignment in the finite-width regime (shown in (c)), enhancing generalization. 
}
\end{figure}

We consider a transformer-like \citep{vaswani2017attention}
architecture consisting of a linear input projection layer; $L$
multi-head self-attention (MHA) layers, each having $H$ attention heads;
and a linear readout layer. The network input $x\in\mathbb{R}^{N_{0}\times T}$
is a sequence of $T$ tokens $x_{t}\in\mathbb{R}^{N_{0}}$,
with token index $t \in \{1,\ldots,T\}$, and dimension $N_{0}$.
The input projection layer performs the transformation
\begin{equation}
x_{t}^{\left(1\right)}=\frac{1}{\sqrt{N_{0}}}V^{\left(0\right)}\cdot x_{t},\qquad\qquad V^{\left(0\right)}\in\mathbb{R}^{N\times N_{0}},\label{eq:input_projection}
\end{equation}
where $N$ is the hidden layers' width. With the operator "$\cdot$" we denote matrix-matrix or matrix-vector multiplication. The $\ell$-th MHA layer with index $\ell \in \{1,\ldots,L\}$
performs the transformation
\begin{equation}
x_{t}^{\left(\ell+1\right)}=\frac{1}{\sqrt{NH}}\sum_{h=1}^{H}\sum_{s=1}^{T}V^{\left(\ell\right)h}\cdot x_{s}^{\left(\ell\right)}\Omega_{st}^{\left(\ell\right)h},\qquad\qquad V^{\left(\ell\right)h}\in\mathbb{R}^{N\times N},\label{eq:MHA_layer_1}
\end{equation}
\Fnotes{We should find a consistent matrix product notation. Proposal: no symbol to indicate matrix product + use transpose for scalar product, so that dimensions definitions are all clear.}
where, for each head $h$, we define the attention matrix $\Omega^{\left(\ell\right)h}\in\mathbb{R}^{T\times T}$
with matrix elements
\begin{equation}
\Omega_{st}^{\left(\ell\right)h}=\zeta\left(\frac{1}{N_{0}\sqrt{G}}x_{s}^{\top}\cdot W_{K}^{\left(\ell\right)h\top}\cdot W_{Q}^{\left(\ell\right)h}\cdot x_{t}\right),\qquad\qquad W_{Q}^{\left(\ell\right)h},W_{K}^{\left(\ell\right)h}\in\mathbb{R}^{G\times N_{0}}.\label{eq:attention_matrix-1}
\end{equation}
Here $\zeta$ is the softmax function, applied along the direction
of the token index $s$, while $G$ is the dimension of the query-key feature space. The
linear readout returns the scalar output
\begin{equation}
f=\frac{1}{\sqrt{N}}a\cdot x_{t^*}^{\left(L+1\right)},\qquad\qquad a\in\mathbb{R}^{1\times N}.\label{eq:linear_readout}
\end{equation}
\Lnotes{I have changed the definition of a to just be a N-dim vector, rather than 1xN. This unnecessarily makes the notation heavy. I have also removed the traspose in the order param interpretation, as a consequence.}Here $x_{t^*}$ can stand for different options for reducing the token dimension
at readout, namely reading from a specific token $t^{*}$
or averaging over all tokens ($x_{t^{*}}\coloneqq\frac{1}{T}\sum x_{t}$).
The network's learnable parameters are the input projection weights
$V^{\left(0\right)}$; the value, query and key weights $\left\{ V^{\left(\ell\right)h},W_{Q}^{\left(\ell\right)h},W_{K}^{\left(\ell\right)h}\right\} _{\ell,h=1}^{L,H}$;
and the readout weights $a$.

\textbf{Comparison to the Standard Transformer.}
The above architecture presents two main simplifications w.r.t.~the standard transformer. First, the network is linear in the value
weights, while the standard transformer has a nonlinear feedforward block
after each MHA layer. Second, in any layer $\ell$, the attention
(Eq.~\ref{eq:attention_matrix-1}) is always computed as a direct function of the
bare input $x$, rather than the processed input
$x^{\left(\ell\right)}$.
These simplifications allow us to apply back-propagating kernel renormalization (BPKR) techniques \cite{li2021statistical,li2022globally}, enabling the characterization of the network beyond the GP limit.
Despite these simplifications, the insights gained by
going beyond the GP limit are substantial: we will show that, in the
finite-width regime, an important mechanism---\textit{task-relevant kernel
combination}---emerges, accounting for a considerable improvement
in generalization performance.

\textbf{Attention paths formulation.} Note that, despite the linearization in the value weights, the network
is still highly nonlinear in the input, thanks to the attention operation
(Eq.~\ref{eq:attention_matrix-1}). This can be seen by the following
equivalent description of the network (\Figref{fig:scheme}(a)). We introduce the concept of
\textit{attention paths}, by defining a path ``index'' $\pi\coloneqq (h_{1},h_{2},\ldots,h_{L})$, where $h_{1},\ldots,h_{L}\in\left\{ 1,\ldots,H\right\} $, which
uniquely identifies each possible combination of the head indices across layers,
i.e., each possible path through the attention heads. The network output
can be rewritten as
\begin{equation}
f=\frac{1}{\sqrt{H^{L}NN_{0}}}\sum_{\pi\in\Pi}V^{\left(\mathrm{eff}\right)\pi}\cdot V^{\left(0\right)}\cdot\xi^{\pi}\label{eq:effective_description-MAIN}
\end{equation}
where $\Pi$ is the set of all possible paths, and we define the ``\textit{effective weights}'' as
\begin{equation}
V^{\left(\mathrm{eff}\right)\pi}\coloneqq\frac{1}{\sqrt{N^{L}}} a\cdot V^{\left(L\right)h_{L}}\cdot V^{\left(L-1\right)h_{L-1}}\cdot\ldots\cdot V^{\left(2\right)h_{2}}\cdot V^{\left(1\right)h_{1}},\quad\quad V^{\left(\mathrm{eff}\right)\pi}\in\mathbb{R}^{1\times  N}\label{eq:V_effective-MAIN}
\end{equation}
\Lnotes{changed the definition of a (see above), consequently also Veff is a N-dim vector, rather than 1xN-dim vector. This was only making the notation heavier for no particular reason.}
and the ``\textit{attentioned input}'' as
\begin{equation}
\xi^{\pi}\coloneqq \sum_{t_{0},\ldots,t_{L-1}=1}^{T}x_{t_{0}}\Omega_{t_{0}t_{1}}^{\left(1\right)h_{1}}\Omega_{t_{1}t_{2}}^{\left(2\right)h_{2}}\ldots\Omega_{t_{L-2}t_{L-1}}^{\left(L-1\right)h_{L-1}}\Omega_{t_{L-1}t^{*}}^{\left(L\right)h_{L}}\,,\quad\quad \xi^{\pi}\in\mathbb{R}^{  N_0}.\label{eq:attentioned_input-MAIN}
\end{equation}
In Eq.~(\ref{eq:effective_description-MAIN}), the network can be seen as a
deep linear network applied to a nonlinearly expanded input---the
attentioned input $\xi^{\pi}$. Through Eq.~(\ref{eq:attentioned_input-MAIN}), we
can see that the bare input $x$ is nonlinearly expanded from
an $N_{0}$-dimensional space to an $N_{0}H^{L}$-dimensional space, by means of $H^{L}$ nonlinear operations: one for each attention path. 

The goal of our theory is to understand how the network learns to
combine these different attention paths, by means of the effective
weights (Eq.~\ref{eq:V_effective-MAIN}).
Note that the network also has other learnable parameters:
the query and key weights, which parameterize the nonlinear expansion of the input to $\xi$.
The learning of these parameters is not described by our theory.
As we will see in Sec.~\ref{sec:Theory}, our theory characterizes the learned effective weights (Eq.~\ref{eq:V_effective-MAIN}) for a given, fixed realization of the query and key weights.

\section{Theory\label{sec:Theory}}

A fundamental quest of deep learning theory is to understand how deep
neural networks, which are often overparameterized, manage to avoid
overfitting, achieving good generalization performance \citep{belkin2019reconciling, zhang2017understanding}. One important
role is played by the specific choice of network architecture, which
can impose an inductive bias towards better generalizing configurations
of parameters, among the many that fit the training data. To study this problem, we adopt the Bayesian framework. Given a dataset of $P$ example-label pairs $\left\{ x^{\mu},y^{\mu}\right\} _{\mu=1}^{P}$, we seek to characterize the Bayesian posterior \citep{tishby1989consistent, mackay1992practical, neal2012bayesian}, or \textit{Gibbs distribution},  
over the parameters $\Theta\coloneqq\left(V^{\left(0\right)},\left\{ V^{\left(\ell\right)h}\right\} _{\ell,h=1}^{L,H},a\right)$

\begin{equation}
p\left(\Theta\right)\propto\exp\left\{ -\frac{1}{2\mathcal{T}}\sum_{\mu=1}^{P}\left[f\left(x^{\mu},\Theta\right)-y^{\mu}\right]^{2}-\frac{1}{2\sigma^{2}}\left\Vert \Theta\right\Vert ^{2}\right\}\, . \label{eq:Gibbs_distribution}
\end{equation}
Here $f\left(x^{\mu},\Theta\right)$ is the network output (Eq.~\ref{eq:linear_readout}) corresponding to the input $x^{\mu}$, where we emphasize its dependence on $\Theta$, $\left\Vert \cdot\right\Vert $ is the Frobenius norm, $\sigma^{2}$ is the variance of the weights' Gaussian prior
(set to $\sigma=1$ throughout this paper), and $\mathcal{T}> 0$ is the error variance, or  \textit{Gibbs temperature} (not to be confused with the number of tokens
$T$). Characterizing the Gibbs distribution allows to gain insights into the inductive bias imposed by the network architecture. Indeed, note that, in overparameterized networks, the Gibbs distribution for $\mathcal{T}\to0^{+}$ describes the statistics
of those parameter configurations that perfectly fit the training
data, with a bias towards small weights induced by the Gaussian prior. These statistics depend on the choice of network architecture, which can therefore bias the distribution towards better generalizing parameter configurations. 
For $\mathcal{T}>0$, parameter configurations that do not achieve
perfect fitting are also allowed, which can help to prevent overfitting.

Note that, as discussed at the end of Sec.~\ref{sec:Model}, we characterize
the statistics of the weights $\Theta$ (the linear projection, value, and readout weights) for a fixed realization of
the query and key weights. The fixed query and key weights can be given, for example, by pre-training the network with gradient descent, or by some task-informed initialization. In Sec.~\ref{subsec:One-shot-image-classification}
we will show that the insights gained by our theory on the weights
$\Theta$ can also be applied to the network trained with gradient descent on all of its learnable parameters, including the query and key weights. 

The main theoretical result of this work is an expression for the expectation $\mathbb{E}\left[f\left(x^{*}\right)\right]$ of the network's output on a new test example $x^{*}$, under the Gibbs distribution (Eq.~\ref{eq:Gibbs_distribution}). In Sec.~\ref{sec:statement} below, we provide this formal result, accompanied by a sketch of its derivation and a discussion of the significance of the infinite-dimensional---or thermodynamic--- limit under which our result is derived.
In Sec.~\ref{sec:implications} we discuss the result's interpretation and its insights into the network's generalization capabilities.

\subsection{Statement of theoretical results\label{sec:statement}}
\textbf{Definitions}. Consider a training dataset consisting of $P$ inputs $x^{\mu}\in\mathbb{R}^{N_{0}\times T}$ and associated labels $y^{\mu}\in\mathbb{R}$, where $\mu=1,\ldots P$. Call $X\coloneqq\{x^{\mu}\}_{\mu=1}^{P}$ the set of training inputs and $Y\in\mathbb{R}^{P}$ the vector of training labels with $\mu$-th component $y^{\mu}$. Consider a network defined by Eqs.~(\ref{eq:input_projection}-\ref{eq:linear_readout}) and in particular call $f^{*}$ the network output (Eq.~\ref{eq:linear_readout}) corresponding to a test input $x^{*}\in\mathbb{R}^{N_{0}\times T}$.

\textbf{Assumptions}. Assume the query and key weights $\left\{ W_{Q}^{\left(\ell\right)h},W_{K}^{\left(\ell\right)h}\right\} _{\ell,h=1}^{L,H}$ are fixed, while all other weights $\ensuremath{\Theta\coloneqq\left(V^{\left(0\right)},\left\{ V^{\left(\ell\right)h}\right\} _{\ell,h=1}^{L,H},a\right)}$ are distributed according to the Bayesian posterior distribution defined in Eq.~(\ref{eq:Gibbs_distribution}). Assume the “thermodynamic limit” $N,N_{0},P\to\infty$, with $P/N\coloneqq\alpha\in\mathbb{R}^{+}$ and $P/(N_{0}H^{L})\coloneqq\alpha_{0}\in\mathbb{R}^{+}$, where $\alpha$, $\alpha_{0}$ as well as other size parameters $T,H,L\in\mathbb{\mathbb{N}}$ are finite.

\textbf{Result 1}. The mean predictor under the posterior distribution (Eq.~\ref{eq:Gibbs_distribution}) is given by

\begin{equation}
\mathbb{E}\left[f^{*}\right]=k^{\top}\cdot\left(K+\mathcal{T}\mathbb{I}\right)^{-1}Y,\label{eq:mean_predictor}
\end{equation}

The vector $k\in\mathbb{R}^{P\times1}$ and the matrix $K\in\mathbb{R}^{P\times P}$, called training kernel, are defined in terms of a kernel function $\mathcal{K}:\mathbb{R}^{N_{0}\times T}\times\mathbb{R}^{N_{0}\times T}\to\mathbb{R}$ as $k^{\mu}\coloneqq\mathcal{K}\left(x^{*},x^{\mu}\right)$ and $K^{\mu\nu}\coloneqq\mathcal{K}\left(x^{\mu},x^{\nu}\right)$, for $\mu,\nu=1,\dots,P$. The kernel function is given by 

\begin{equation}
\mathcal{K}\left(x,x'\right)=\frac{1}{H^{L}}\sum_{\pi,\pi'\in\Pi}U^{\pi\pi'}C_{\pi\pi'}\qquad\mathrm{with}\qquad C_{\pi\pi'}\coloneqq\frac{1}{N_{0}}\xi^{\pi}\left(x\right)^{\top}\cdot\xi^{\pi'}\left(x'\right)\,,\label{eq:Kernel}
\end{equation}

where $\xi^{\pi}\left(x\right)$ is the “attentioned input” corresponding to an input $x\in\mathbb{R}^{N_{0}\times T}$, along path $\pi\in\Pi$, as defined in Eq.~(\ref{eq:attentioned_input-MAIN}). The kernel function depends on a positive semi-definite matrix $U\in\mathbb{R}^{H^{L}\times H^{L}}$, called \textit{order parameter}, which is given by

\begin{equation}
U=\underset{{\tilde{U}}}{\mathrm{argmin}}\,S(\tilde{U};X,Y) \qquad\mathrm{with}\qquad S(U;X,Y)=-\mathcal{L}(U)+\alpha\mathcal{E}(U;X,Y)\,,\label{eq:argmin} 
\end{equation}

The scalar function $S$, called the \textit{action}, consists of an “\textit{entropy}” term $\mathcal{L}$, and an “\textit{energy}” term

\begin{equation}
    \mathcal{E}(U;X,Y)=\frac{1}{P}\ln\det\left(K(U;X)+\mathcal{T}\mathbb{I}\right)+\frac{1}{P}Y^{\top}\cdot\left(K(U;X)+\mathcal{T}\mathbb{I}\right)^{-1}\cdot Y,\label{eq:energy}
\end{equation}

where $K(U;X)\coloneqq K$ is the training kernel matrix. The expression for the entropy $\mathcal{L}$ is lengthy and is given in Appendix~\ref{subsec:Predictor-statistics}. In the special case of $H=1$, $U$ is a scalar, and $\mathcal{L}\left(U\right)=-\sigma^{-2\left(L+1\right)}U+\ln\left(U\right)$. For general $H$, the entropy $\mathcal{L}\left(U\right)$ is always maximized by $\ensuremath{U^{\pi\pi'}=\sigma^{2\left(L+1\right)}\delta_{\pi,\pi'}}$, which therefore is the solution of Eq.~(\ref{eq:argmin}) in the GP limit defined by $\alpha\to0^{+}$.

\textbf{Result 2}. The matrix \textit{U} obeys the following relation

\begin{equation}
    U^{\pi\pi'}=\frac{1}{N}\mathbb{E}[V^{\left(\mathrm{eff}\right)\pi}\cdot V^{\left(\mathrm{eff}\right)\pi'\top}]\,\label{eq:o_param_intepreatation}
\end{equation}

where $V^{\left(\mathrm{eff}\right)\pi}\in\mathbb{R}^{1\times N}$ are the effective weights along path $\pi$ , defined in Eq.~(\ref{eq:V_effective-MAIN}).

\textbf{Derivation.} See Appendix~\ref{sec:appendix-theory}. 

The derivation, which uses the BPKR technique \cite{li2021statistical,li2022globally}, can be sketched as follows. Computing $\mathbb{E}\left[f^{*}\right]$ under the posterior distribution $p\left(\Theta\right)$ involves evaluating a high-dimensional integral in the weights $\Theta$. The idea is to first reduce this computation into an integration over a lower-dimensional, `macroscopic' variable $U$. Importantly, while $\Theta$  becomes infinite-dimensional as $N\to\infty$, $U$ remains finite-dimensional. The reduced integral is an expectation of the r.h.s.~of Eq.~(\ref{eq:mean_predictor}), treated as a function of $U$, under the distribution $p\left(U\right)\propto\exp\left\{-\frac{1}{2}NS\left(U\right)\right\}$, where $S$ is the action defined in Eq.~(\ref{eq:argmin}). Then, this integral can be solved in the thermodynamic limit $N\to\infty$, using the saddle-point method, which implies evaluating Eq.~(\ref{eq:mean_predictor}) at the $U$ that minimizes the action (cf. Eq.~\ref{eq:argmin}). Crucially, the end result is fully characterized by this low-dimensional quantity $U$, commonly called \emph{order parameter} in physics, which has a direct interpretation in terms of the network weights, 
given by Eq.~(\ref{eq:o_param_intepreatation}).

In practice, the results obtained in the thermodynamic limit represent a good approximation also for the case of large but finite $N$. In this regard, the scaling of other hyperparameters with $N\to\infty$  is of particular importance, especially the number of training examples $P$. In the GP limit, one considers $P$ finite. This is also called the \textit{infinite-width} thermodynamic limit because in practice, for a given and typically large $P$, it is a good approximation only for very wide networks, when $N\gg P$. In contrast, here we consider the \textit{finite-width} limit in which $P/N=\alpha\in\mathbb{R}^{+}$ (which includes the GP limit for $\alpha\to0^+$). As can be seen from Eq.~(\ref{eq:argmin}), the action gains a new term for $\alpha>0$, which, as we shall discuss below, is fundamental to account for the learning of an attention paths interplay. Finally, we note that in our numerical experiments (Sec.~\ref{Experiments}) we will consider Bayesian networks which are overparameterized, i.e. $P<N_0 H^L$, which is the network capacity at fixed query and key weights.

\subsection{Results interpretation and implications for generalization capability\label{sec:implications}}

Eq.~(\ref{eq:mean_predictor}) is a commonly found expression in thermodynamic theories of Bayesian learning, relating the network's mean predictor to kernel regression. In particular, the theory of kernel regression \citep{canatar2021spectral} suggests that generalization improves when the training kernel $K$ is well aligned with the task, meaning its largest principal components (PCs) are well aligned with the vector of training labels $Y$.

Our result for the transformer's kernel (Eq.~\ref{eq:Kernel}) enables insights into how the transformer architecture favors this kernel-task alignment (\Figref{fig:scheme}(d)). The kernel consists of the sum, weighted by the order parameter $U$, of many \textit{path-path kernels} \Fnotes{what about ``path-coupling kernels" .. I do not like ``path-path" too much}
$C_{\pi\pi'}$, each computing the similarity between the attentioned
input on two attention paths $\pi$ and $\pi'$. A notable property of the
multi-head architecture is that, despite the number of attention heads
growing only linearly with the depth $L$, the number of attention
paths grows exponentially $\propto H^{L}$. Therefore, the network
has at its disposal an exponentially large number of path-path kernels,
which it can learn, through $U$, to optimally combine into a total
kernel with improved task alignment.

This phenomenon, which we term \textit{task-relevant kernel combination}, is indeed predicted by our results Eqs.~(\ref{eq:argmin}-\ref{eq:energy}). These state that the learned $U$ minimizes a function $S$ (Eq.~\ref{eq:argmin}), which, through the energy term $\mathcal{E}$, favors kernel-task alignment. This can be seen by interpreting the energy term (Eq.~\ref{eq:energy}) as the negative log-likelihood of the training labels $Y$ under a centered Gaussian distribution, whose covariance matrix is the training kernel $K$. This negative log-likelihood can be minimized by aligning the largest PCs of the covariance (i.e. the kernel $K$) as much as possible with $Y$ (\Figref{fig:scheme}(c)).

In contrast, in the GP limit $\alpha\to0^+$, the action $S$ (Eq.~\ref{eq:argmin}) consists only of the entropy term $\mathcal{L}$, which does not contain any task relevant information. 
Its only effect is  to attract $U$ towards the GP limit solution $U^{\pi\pi'}=\sigma^{2\left(L+1\right)}\delta_{\pi,\pi'}$.
Note that, in this limit, the benefits of kernel combination \Knotes{we can not use the terminology ``kernel combination'' anymore right?} are lost (\Figref{fig:scheme}(d), bottom line):
First, out of all the path-path kernels $C_{\pi\pi'}$, only the \textit{same-path
kernels} ($\pi=\pi'$) are used, while the \textit{cross-path kernels}
($\pi\neq\pi'$) are discarded; Second, all same-path kernels are
weighted equally, without making use of any task-specific information.
Note that this is true not only for our simplified model, but also
for the full transformer architecture under its known GP limit \citep{hron2020infinite}. A task-relevant kernel combination can therefore only emerge beyond the GP limit, in the finite-width regime $\alpha>0$ studied in this work.

Finally, our result Eq.~(\ref{eq:o_param_intepreatation}) relates the order parameter to a macroscopic measure of the network weights, allowing for a direct interpretation of the
kernel combination mechanism: correlating the effective weights across paths allows the
network to make use of cross-path kernels, while controlling their
magnitude allows to weigh the different path-path kernels in a task-relevant
manner.

\section{Experiments\label{Experiments}}

To corroborate our theoretical results, we “train” our model (Eqs.~\ref{eq:input_projection}-\ref{eq:linear_readout}) by sampling its weights $\Theta$  (i.e. all weights except the fixed query and key weights) from the posterior distribution Eq.~(\ref{eq:Gibbs_distribution}), using Hamiltonian Monte Carlo sampling (see Appendix~\ref{appendix_Hamiltonian_Montecarlo} for details). We consider the following two tasks:
hidden Markov chain (HMC) classification,
and one-shot image classification by in-context learning. The first
task is defined on a synthetic dataset. Its purpose is to have a
minimal, controllable setting to illustrate the effects of
task-relevant kernel combination.
In the second task, we will proceed to show analogous effects on classic image datasets (Omniglot \citep{omniglot}, MNIST \citep{lecun1998mnist}, and FashionMNIST \citep{xiao2017fashion}), and compare these results with those obtained from the same network trained with standard gradient descent on all of its parameters (i.e. including the query and key weights).

\subsection{Hidden Markov chain sequence classification\label{subsec:Hidden-Markov-chain}}

\begin{figure}
\centering
\includegraphics[width=0.95\textwidth]{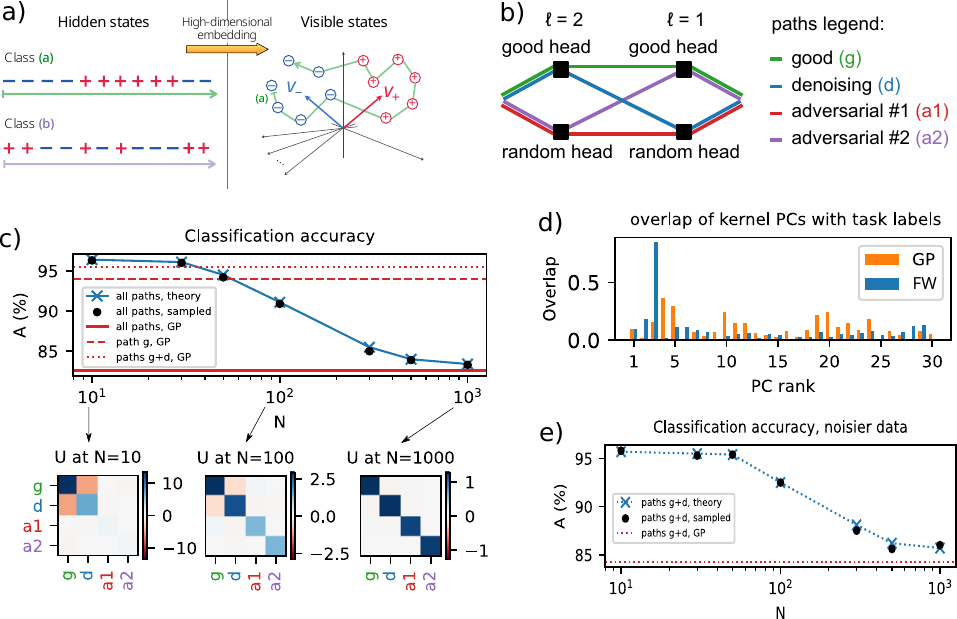}
\caption{\label{fig:markov}{\bf Hidden Markov chain task.} {\bf (a)} Illustration of the task.
{\bf (b)} Schematics of the network and its attention paths. {\bf (c) Top:} Classification
accuracy for varying $N$ (theory: blue crosses, joined by blue line;
samples: black dots). Red lines: GP limit for a network consisting
of all paths (solid), the good path (dashed), and the good and denoising
paths (dotted). {\bf Bottom:} Matrix
elements of $U$, for varying $N$. The matrix indices are labeled
with the corresponding path name, according to the legend in (b). {\bf (d)} Normalized overlap, or cosine similarity, between
the PCs of the kernel $K$ and the vector of task labels $Y$ ($N=10$:
blue; GP limit: orange). PCs are ranked by their eigenvalues, from
largest to smallest. Only the first $30$ PCs are shown.
{\bf (e)} Same as (c), but for increased $\sigma_{\perp}=5$ and a network
consisting of only the good and denoising paths.}
\end{figure}

\textbf{Task definition.} The HMC classification task is defined as follows (\Figref{fig:markov}(a)).
The $\mu$-th example in the dataset corresponds to an hidden Markov chain $q_{1}^{\mu},\ldots,q_{T}^{\mu}$
of length $T=30$\Fnotes{add this notation to the figure}, alternating between two hidden states, $q_{t}^{\mu}\in\left\{ +,-\right\} $.
The probability of transition  to the opposite state ($\pm\to\mp$) is $p^{\mu}$. The $\mu$-th chain can belong to one of two classes, labeled $y^{\mu}=\pm1$, depending on whether $p^{\mu}=0.3$ or $p^{\mu}=0.7$, respectively.
The input tokens are a noisy, higher dimensional
representation of the hidden states. These are given by $x_{t}^{\mu}=v_{q_{t}^{\mu}}+\eta_{t}^{\mu}$,
where $v_{\pm}\in\mathbb{R}^{N_{0}}$ are two orthogonal feature vectors
corresponding to the states ``$\pm$'', with $\mathcal{O}\left(1\right)$
entries, while $\eta_{t}^{\mu}$ is a zero-mean Gaussian noise, with
$\langle\eta_{t}^{\mu}\eta_{t'}^{\mu'\top}\rangle=\delta_{\mu,\mu'}\delta_{t,t'}(\sigma_{\parallel}^{2}P_{\parallel}^{\top}\cdot P_{\parallel}+\sigma_{\perp}^{2}P_{\perp}^{\top}\cdot P_{\perp})$,
where $P_{\parallel}$ and $P_{\perp}$ are the projectors along the
subspace parallel or perpendicular to the plane spanned by $v_{+}$
and $v_{-}$. Unless specified, $\sigma_{\parallel}=\sigma_{\perp}=1$.
The separate parameterization of the parallel ($\sigma_{\parallel}$)
and perpendicular ($\sigma_{\perp}$) noise strengths is motivated
by their distinct effect on task performance: while the first corrupts
information about the underlying hidden states, inevitably putting
an upper bound on the classification accuracy, the second can always
be filtered out by learning appropriate weights. 
We use $P=100$ examples for training.
We test the network performance in terms of the classification accuracy
$A=\frac{1}{P^{*}}\sum_{\mu}\delta_{y^{\mu},\mathrm{sign}\left(\left\langle f^{\mu}\right\rangle \right)}$,
where the sum is over a number $P^{*}=1000$ of test examples. Additional task details are given in Appendix~\ref{sec:markov_def_appendix}.

\subsubsection{Results}

We consider a network of $L=2$ layers and $H=2$ heads per layer,
with readout from the first token. The network has a total
of $4$ attention paths, schematically depicted in \Figref{fig:markov}(b).
For this synthetic task, we design the fixed query and key weights, and therefore the network's attention paths, to clearly
illustrate the effects of task-relevant kernel combination (for details, see Appendix~\ref{apdx_query-key-init}).

We design the first head of each layer to give rise to a ``good'' attention path (green path) such that a network consisting of
this good path alone achieves a high classification accuracy, $A\sim94\%$. Along this path, the first head makes use of the Markov nature of the task by attending exclusively to nearby tokens, and only if they correspond to the same hidden state $\pm$; the second head performs uniform attention, effectively counting how many times the first head detected the same-state transition $\pm\to\pm$. In contrast, each layer's second head is initialized randomly. This results in the three remaining paths having chance-level
classification accuracy $A\sim50\%$, when considered in isolation.
However, these paths have very different effects, when combined with
the good path. We term two of these paths ``adversarial'' (red and
purple paths) because they deteriorate the network performance, while
we term the remaining path ``denoising'' (blue path) because it
can be effectively combined with the good path to improve robustness
to noisy data.

In \Figref{fig:markov}(c, top) we show the network's classification accuracy
as a function of the width $N$ (blue, solid curve), compared to the
GP limit (red, solid line). At lower $N$, well into
the finite-width regime, we observe a considerable improvement in performance with respect to the GP limit. This can be understood in terms of an improved kernel-task alignment, as shown in \Figref{fig:markov}(d). 

This improved alignment is ensured by the order parameter $U$, plotted
in \Figref{fig:markov}(c, bottom) for varying $N$. For $N=10$, well into the
finite-width regime, the order parameter clearly implements the two
main benefits of kernel combination: the possibility to weigh the
path-path kernels differently, and the ability to make use of the
cross-path kernels. The first benefit is particularly apparent in
the suppression of all kernels associated with the adversarial paths.
In contrast, when $N=1000$ and the order parameter is very close
to its GP limit  $U^{\pi\pi'}=\delta_{\pi,\pi'}$, these paths
are not suppressed, causing a deterioration in performance compared
to that of the good path alone (red, dashed line in \Figref{fig:markov}(c, top)).
The second benefit is apparent in the strong off-diagonals of $U$,
anti-correlating the good and denoising paths. We can see that, while
also in the GP limit the denoising and good paths combined (dotted,
red line in \Figref{fig:markov}(c, top)) have a better performance than the
good path alone (dashed, red line), the performance boost is even
higher in the renormalized regime, which makes use of the cross-path
kernels. This additional improvement in performance becomes more apparent
with noisier data. This is shown in \Figref{fig:markov}(e), where we
plot the classification accuracy of the network consisting of only
the good and denoising paths, on data with stronger
perpendicular noise $\sigma_{\perp}=5$.

\subsection{One-shot image classification\label{subsec:One-shot-image-classification}}

\begin{figure}
\centering
\includegraphics[width=.95\textwidth]{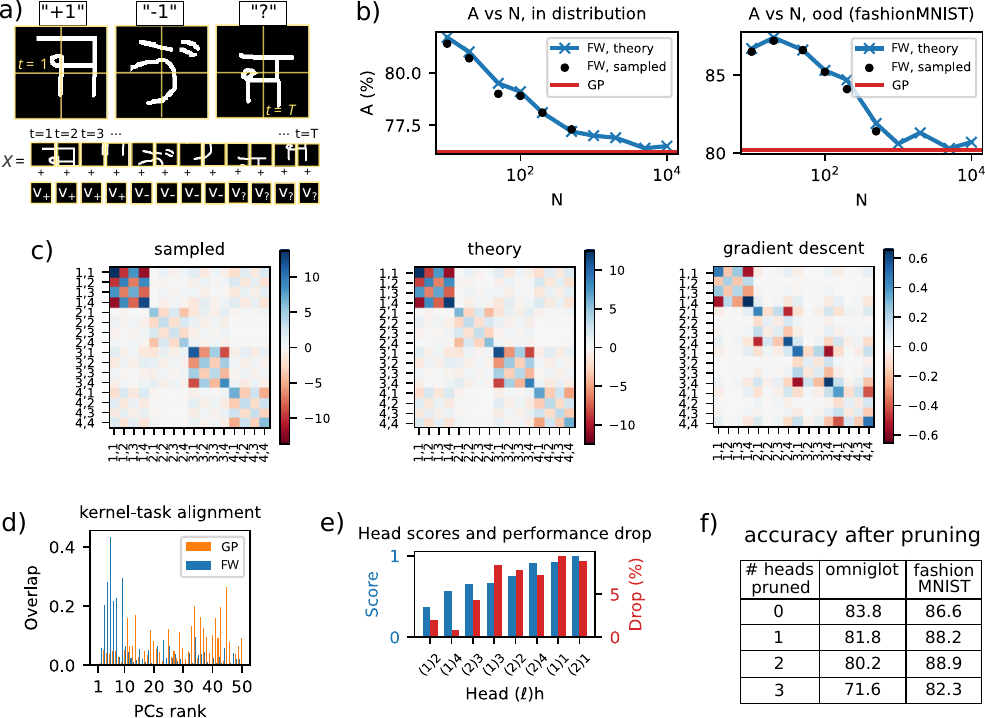}
\caption{\label{fig:one_shot}\textbf{One-shot image classification task.} \textbf{(a)} Scheme
of the task. \textbf{(b)} Classification accuracy in the GP limit (red line)
and the finite-width regime (FW) for varying $N$ (theory: blue crosses,
joined by blue line; samples: black dots).
\textbf{(c)} Matrix elements of
$U$. The ``theory'' and ``sampled'' $U$s are for $N=10$. The
matrix indices are labeled with the path index $\pi=(h_{1},h_{2})$.
\textbf{(d)} Kernel PCs' overlap
with the task, in the GP limit and in the finite-width regime for
$N=10$. Only the first $50$ PCs are shown.
\textbf{(e)} Head score (blue) and performance drop (red) after pruning the
head, for the model trained with gradient descent. \textbf{(f)} Classification
accuracy of the model trained with gradient descent, after pruning
a growing number of heads, in order of their head score.}
\end{figure}

\textbf{Task definition.} The one-shot image classification task (\Figref{fig:one_shot}(a)) is formulated in an in-context learning setting. The network is presented with a sequence of three image-label
pairs. The first two images belong to two distinct classes of a categorized
dataset (Omniglot, FashionMNIST or MNIST in our case). They are assigned
the label ``$+$'' or ``$-$'' in no particular order. The third
image is assigned the label ``$?$'', and belongs to one of the
classes of the first two images. The network has to output $\pm1$
according to the label of the matching image. The sequence is fed
to the network as follows. Following the idea of the vision transformer
(ViT) \citep{dosovitskiy2020image}, each image is divided into $p$ patches. The patch $i \in \{1,\ldots,p\}$
of image $a \in \{1,2,3\}$ corresponds to the token $x_{(a-1)p+i}$, for
a total of $T=3p$ tokens.
We encode the labels $+$, $-$, $?$ using three fixed random vectors $v_{+}, v_{-}, v_{?} \in \mathbb{R}^{N_{0}}$,
which we directly add to each patch (i.e., token) of the corresponding image.
We also encode the token position with
additive sinusoidal positional encoding \citep{vaswani2017attention}. The network is trained on the Omniglot dataset \citep{omniglot}, while we test its classification accuracy on both in-distribution (ID) unseen classes of Omniglot, and out-of-distribution (OOD) FashionMNIST dataset (we also report results on MNIST in Appendix~\ref{sec:One-shot-appendix}).

\subsubsection{Results}

We consider a network of $L=2$ attention layers and $H=4$ heads
per layer, with average pooling readout, trained on a subset of $P=600$ examples from Omniglot (analogous results for a deeper network with $L=3$, $H=3$ are also reported in Appendix~\ref{sec:class_accuracy}).
For the fixed query and key weights required by our Bayesian network, we use the query and key weights obtained from training the same network using gradient descent, with $N=512$, $G=128$,
and $P=528k$ (i.e., the entire training set from Omniglot).
We refer to Appendix~\ref{sec:grad-descent-appendix} for further details on this process.

The plots shown in \Figref{fig:one_shot} are analogous to those for the
HMC task (\Figref{fig:markov}), and illustrate analogous kernel combination
phenomena. \Figref{fig:one_shot}(b) shows the classification
accuracy for varying $N$. Again, we observe a performance gap between
the finite-width and GP regimes. Interestingly, this improvement in
performance is preserved also OOD, on FashionMNIST. %
Again, \Figref{fig:one_shot}(d) shows that the performance gap can be understood in terms of an improved
kernel-task alignment: PCs that are well aligned with $Y$ are of
higher rank, and have a larger overlap than in the GP limit. 

The order parameter (\Figref{fig:one_shot}(c), ``theory'' and ``sampled'')
for $N=10$ is clearly far from its GP limit, accounting for the improvement
in performance observed in the finite-width regime. We observe similar
kernel combination phenomena as in the HMC task, with strong off-diagonal
elements, and a stronger weighting of certain paths w.r.t. others.
Interestingly, the block diagonal structure of the order parameter allows
for a simple interpretation of the interplay between paths: correlations mostly occur between paths sharing the same head $h_{1}$ in the first layer, which also determines which paths are overall enhanced ($h_{1}=1,3$) or suppressed ($h_{1}=2,4$).

This structure of the order parameter transfers qualitatively well
also to the network trained with gradient descent. In \Figref{fig:one_shot}(c,
``gradient descent'') we show an empirical order parameter, obtained
by computing \Eqref{eq:o_param_intepreatation} using a single realization
of the network's weights trained with gradient descent. Both the order
parameter's block structure and matrix element signs are qualitatively
preserved in this empirical estimate. We emphasize that the network
is trained with the full set of training examples ($P=528k$) rather
than the restricted one used for the Bayesian network ($P=600$), and
on all learnable parameters including the query and key weights, making
this qualitative agreement more relevant to potential applications.
One example application is provided below.


\textbf{Application: model  reduction via head pruning.} Our theory allows us to prune certain heads in the
model trained with gradient descent (leading to a model size and compute reduction), with marginal performance loss.
This is achieved by using the order parameter to assign a score to
each attention head, according to its contribution to kernel combination.
The lowest-scoring heads are then pruned from the model. The head
$h$ at layer $\ell$ is assigned the score $s^{\left(\ell\right)h}=\sum_{\pi,\pi'\in\Pi^{\left(\ell\right)h}}\left|U_{\pi,\pi'}\right|$,
where $\Pi^{\left(\ell\right)h}$ is the set of all paths passing
through that head, and $U$ is the order parameter derived from theory. \Figref{fig:one_shot}(e) shows the score of each head, normalized
by the largest one, compared against the drop in classification accuracy
caused by pruning that head. Note that the network is not retrained after
pruning, but only reevaluated on the test examples. We observe a performance
drop qualitatively in line with the head scores. Most importantly,
the two lowest scoring heads only cause a marginal drop in performance.
In \Figref{fig:one_shot}(f) we report the classification accuracy after
pruning an increasing number of heads, in order of their score. Up
until the first two heads (amounting to $25\%$ of the total number of network parameters), the in-distribution classification accuracy
is only marginally worsen. Interestingly, the OOD classification accuracy
is even improved, possibly indicating an overspecialization of the
pruned heads in solving only the in-distribution task. 
In Appendix \ref{sec:appendix_head_pruning} we achieve an analogous size reduction of $25\%$ on a larger model with $H=8$ heads.

\textbf{Agreement with sampled statistics}. Finally, we note that both figures \ref{fig:markov}(c,e) and \ref{fig:one_shot}(b,c) show good agreement of our
theory with the mean predictor and order parameter sampled from
Eq.~\ref{eq:Gibbs_distribution}. While our theory
becomes exact in the $N,P\to\infty$ limit, the agreement holds even for small $N=10$. In particular, in figures \ref{fig:one_shot}(b,c), it holds even if $N<H^{L}\left(H^{L}+1\right)/2=136$, the number of independent entries in the order parameter, which is supposed to be a finite quantity in our theory.

\section{Conclusion and Discussion\label{sec:discussion}}

\textbf{Conclusion.} We introduce a transformer-like model featuring deep multi-head self-attention, amenable to theoretical characterization within the Bayesian framework. Our results unveil the important role of attention paths in accounting for transformers' remarkable performance. We demonstrate that, in scenarios involving the interplay of attention paths at finite widths, generalization consistently improves compared to the GP regime, where such interplay is absent. Our theory explains this paths interplay in terms of a task-relevant kernel combination mechanism, where the network's total kernel results from the sum of many kernels, specific to pairs of paths, and optimally weighted to improve generalization. This mechanism is confirmed by experiments on both synthetic and real-world sequence classification tasks.
More broadly, our results are relevant to the theory of deep learning, as they provide an example of non-scalar kernel renormalization \citep{li2022globally, zhang2024robust} in a widely adopted architecture such as the transformer, illustrating its importance in accounting for network performance. Non-scalar, as opposed to scalar \citep{li2021statistical}, renormalization can affect the network's mean predictor and therefore lead to improved generalization. Our work provides a novel interpretation of its benefits in terms of an optimized kernel-task alignment.

\textbf{Interpretability of our Theory.} We provide interpretability to the kernel combination mechanism, by relating it to observable structures in the network weights, specifically to their magnitude and correlations. These predicted structures transfer well outside the Bayesian framework, to the weights of networks trained with gradient descent, broadening the applicability of our theory. As an example, we show that a trained network can be reduced in size with minimal performance loss, by pruning those heads that are deemed less relevant by our theory. The large size reduction achieved ($25\%$) appears in line with observations that a few specialized heads are responsible for most of the network performance \citep{voita2019analyzing}, and the proposal of head pruning schemes during training \citep{liu23am}. Our theoretical insights may therefore be relevant to the quest for minimalistic and resource-efficient models \citep{touvron2023llama, eldan2023tinystories}. 

\textbf{Limitations and Outlook.} The above results are enabled by our theory's ability to go beyond the GP limit \cite{hron2020infinite}, as well as incorporating a multi-head, multi-layer architecture---a prerequisite for the very existence of attention paths. However, various limitations could still be addressed, opening for exciting research directions. For example, attention in our model is only a function of the bare input, rather than the previous layer's postactivation, as in standard transformers.  In this case, the theoretical challenge would be to disentangle the learning of attention paths interplay from the learning of the attention paths themselves, since now also the attention matrix would depend on the value weights. Another limitation of our model is its linearity in the value weights. It may be possible to heuristically extend the theory to include nonlinear MLP blocks in between attention layers, by replacing the GP path-path kernels appearing in Eq.~(\ref{eq:Kernel}) with the corresponding GP kernels for the nonlinear case---an approach which has proven successful in deep ReLU networks for certain regimes \citep{li2021statistical}. Introducing nonlinearities, strong feature learning may also emerge \cite{geiger2020disentangling,NEURIPS2023_1ec69275,vanmeegen2024codingschemesneuralnetworks}.
Note that, instead, our theory is readily extendable to the case of \textit{linear} MLP blocks, as well as multiple outputs, following  \cite{li2021statistical,li2022globally}. Here we chose a minimal setting focusing only on those renormalization phenomena specific to the transformer architecture. Indeed, the presence of multiple outputs causes the same kind of renormalization independently of the network architecture (i.e., adding two new output indices to the order parameter), while deeper linear blocks would not alter the essence of attention paths interplay, only affecting  details in the entropy part of the action. Extending the theory to include skip connections also seems viable. A very open challenge, instead, is to characterize the learning of the query and key weights, which relates to the more general challenge of extending the BPKR technique to nonlinear deep networks. Finally, our approach characterizes the inductive bias imposed by the network architecture on the parameter configurations that fit the training data, but not the bias imposed by a learning algorithm. It would therefore be interesting to import our theory to methods characterizing deep neural networks' training dynamics \cite{saxe2019mathematical,bordelon2023self,avidan2023connecting}.

\section*{Acknowledgements}
We acknowledge support of the Swartz Foundation,  the Kempner Institute for the Study of Natural and Artificial Intelligence at Harvard University, the Office of Naval Research (ONR) grant No. N0014-23-1-2051, and the Gatsby Charitable Foundation. This research was supported in part by grant NSF PHY-2309135 to the Kavli Institute for Theoretical Physics (KITP). FM was supported by the Simons Foundation (Award Number: 1141576). We have benefitted from helpful discussions with 
Alexander van Meegen, Haozhe Shan and Qianyi Li.

\bibliographystyle{unsrtnat}
\newpage
\appendix
\startcontents[appendices]
\printcontents[appendices]{l}{0}{\part*{Appendices}\setcounter{tocdepth}{2}}

\part{Further discussion on related theory works\label{part:related_works}}
The theoretical properties of attention-based models have been investigated from various perspectives in recent years. Different lines of works have studied the expressivity \cite{fu2024can,edelman2022inductive,hahn2020theoretical,yun2019transformers}, the inductive bias \cite{sahiner2022unraveling,tarzanagh2023max} and the training dynamics \cite{boix2023transformers,li2022theoretical,tian2023scan} of attention layers. In particular, \cite{yun2019transformers} shows that multi-layer attention networks are universal approximators for certain classes of functions, such as equivariant sequence-to-sequence functions, while \cite{hahn2020theoretical} highlights their computational limitations, demonstrating that self-attention cannot model periodic finite-state languages, nor hierarchical structure, unless the number of layers or heads increases with input length. \cite{li2022theoretical,edelman2022inductive} derive error bounds for non-linear models, with trainable queries and keys. As in our work, \cite{geshkovski2023mathematical,fu2024can,jelassi2022vision} focus on the training of value matrices. In \cite{geshkovski2023mathematical,fu2024can} query and key matrices are fixed and untrainable, while \cite{jelassi2022vision} sets them equal to the identity. Several theoretical studies have focused on in-context learning \cite{bai2024transformers,guo2023transformers,li2023transformers,zhang2023trained,zhang2023and}. However, the works mentioned above derive generalization bounds and do not provide a tight characterization of the learning curves. A first tight analysis was done in \cite{rende24}, considering a single-layer model with factored self-attention. The authors of \cite{cui2024phase} provide a tight analysis of a single-layer attention model with trainable tied queries and keys and value weights fixed to the identity. Previous theoretical works have mainly focused on a single transformer block, where attention paths are not defined. For instance, \cite{voita2019analyzing} finds that one transformer block can learn different
linguistic tasks according to the position of the self-attention layer. \cite{li2023transformers} shows that a transformer block can
learn to encode topical models. \cite{jelassi2022vision} shows that one transformer block can encode patch associations. A different line of works has studied attention layers in the infinite-width limit \cite{hron2020infinite,lavie2024towards}. In particular, \cite{hron2020infinite} establishes an equivalence between Gaussian processes and infinitely-wide multi-layer attention models with infinitely many attention heads. \cite{lavie2024towards} leverages this framework and studies the inductive bias of infinitely-wide transformers towards permutation symmetric functions in sequence space.
\newpage
\part{Theory\label{sec:appendix-theory}}

\section{Definitions\label{apdx_definitions}}

We recall the definitions for our transformer model and theory, Sec.~\ref{sec:Model} and \ref{sec:Theory} of the main text. 

\subsection{Variables and parameters}

The hyperparameters are
\begin{align*}
T & :\qquad\text{number of tokens}\\
L & :\qquad\text{number of attention layers}\\
H & :\qquad\text{number of attention heads per layer}\\
N_{0} & :\qquad\text{input width}\\
N & :\qquad\text{hidden layers width}\\
G & :\qquad\text{query-key internal dimension }\\
P & :\qquad\text{number of training examples}\\
\mathcal{T} & :\qquad\text{Gibbs temperature (theory only)}\\
\sigma & :\qquad\text{Gaussian prior variance (theory only)}
\end{align*}

The network weights are
\begin{align*}
V^{\left(0\right)}\in\mathbb{R}^{N\times N_{0}} & :\qquad\text{input projection weights}\\
V^{\left(\ell\right)h}\in\mathbb{R}^{N\times N} & :\qquad\text{value weights }\\
a\in\mathbb{R}^{1\times N} & :\qquad\text{readout weights}\\
W_{Q}^{\left(\ell\right)h},W_{K}^{\left(\ell\right)h}\in\mathbb{R}^{G\times N} & :\qquad\text{\text{query and key weights (fixed in the theory)} }
\end{align*}
where $h=1,\ldots,H$, and $\ell=1,\ldots,L$.

The network input is 
\begin{align*}
x\in\mathbb{R}^{N_{0}\times T} & :\qquad\text{input sequence}\\
x_{t}\in\mathbb{R}^{N_{0}} & :\qquad\text{single token in the sequence}
\end{align*}
where $t=1,\ldots,T$.

We append an additional index $\mu=1,\ldots,P$ to quantities that
refer to a specific example in the training set, such as, e.g.
\begin{align*}
x^{\mu}\in\mathbb{R}^{N_{0}\times T} & :\qquad\text{\ensuremath{\mu}-th example in the training set}\\
y^{\mu}\in\mathbb{R} & :\qquad\text{\ensuremath{\mu}-th training label}
\end{align*}
We further define 
\begin{align*}
X & \coloneqq\left\{ x^{\mu}\right\} _{\mu=1}^{P}\\
Y & \coloneqq\left\{ y^{\mu}\right\} _{\mu=1}^{P}\qquad\text{the vector of training labels}
\end{align*}

\subsection{Model definition}

\begin{figure}
    \centering
    \hspace{-1em}\includegraphics[width=1.1\textwidth]{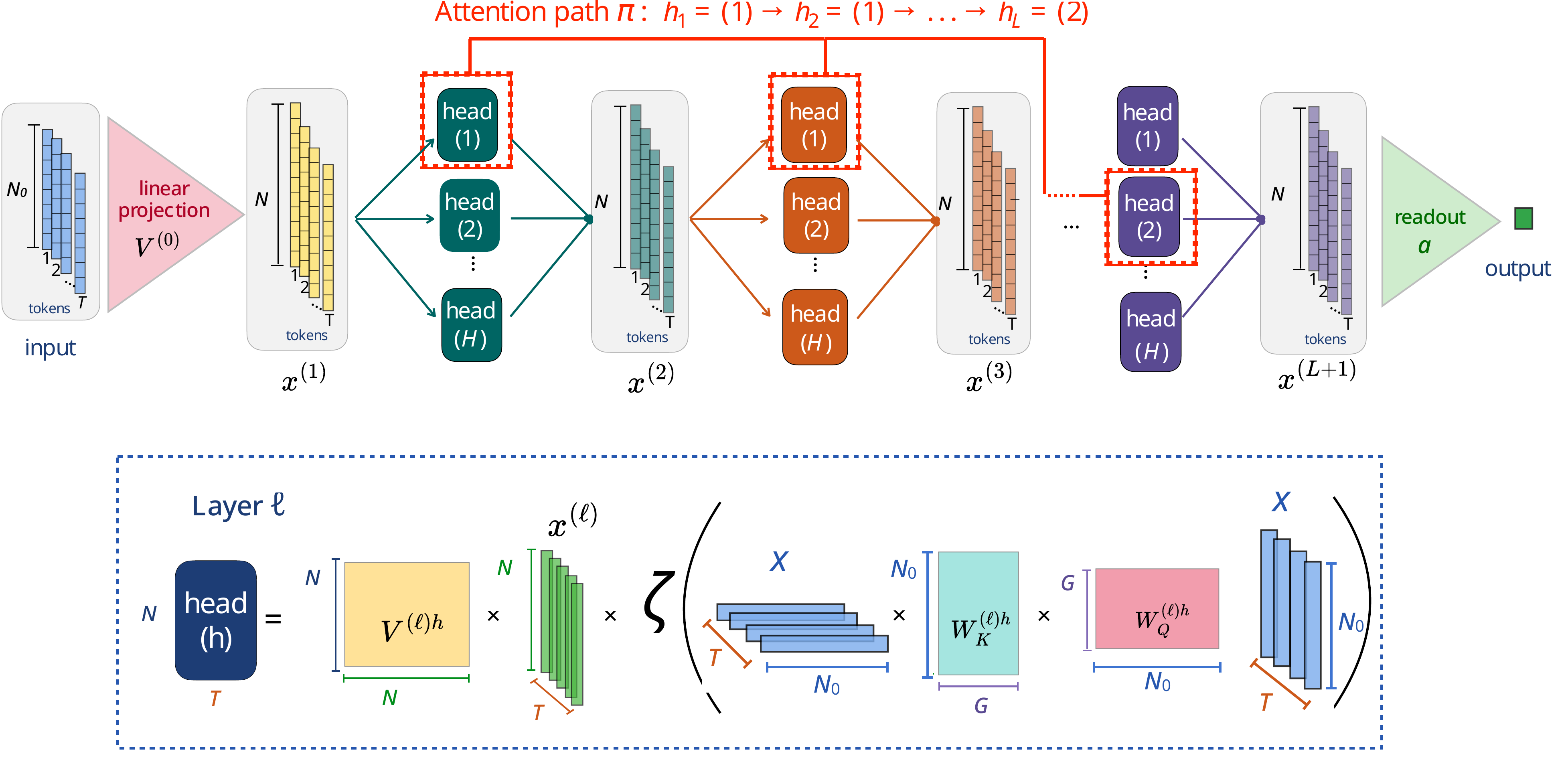}
    \caption{Schematic representation of the architecture under consideration.}
    \label{fig:illustration}
\end{figure}

The network performs the following sequence of input transformations (\Figref{fig:illustration}).
The input projection layer is defined as
\begin{equation}
x_{t}^{\left(1\right)}=\frac{1}{\sqrt{N_{0}}}V^{\left(0\right)}\cdot x_{t}
\end{equation}
The output from each attention layer $\ell=1,\ldots,L$ is defined
recursively as: 
\begin{equation}
x_{t}^{\left(\ell+1\right)}=\frac{1}{\sqrt{NH}}\sum_{h=1}^{H}\sum_{s=1}^{T}V^{\left(\ell\right)h}\cdot x_{s}^{\left(\ell\right)}\Omega_{st}^{\left(\ell\right)h},\qquad t=1,\ldots,T
\end{equation}
where, for each head $h$, we define the attention matrix $\Omega^{\left(\ell\right)h}\in\mathbb{R}^{T\times T}$
with matrix elements 
\begin{equation}
\Omega_{st}^{\left(\ell\right)h}=\zeta\left(\frac{1}{N_{0}\sqrt{G}}x_{s}^{\top}\cdot W_{K}^{\left(\ell\right)h\top}\cdot W_{Q}^{\left(\ell\right)h}\cdot x_{t}\right)\label{eq:attention_matrix-1-1-1}
\end{equation}
The function $\zeta$ denotes the softmax applied along the direction
of the token indexed by $s$. The linear readout is defined as
\begin{equation}
f=\frac{1}{\sqrt{N}}a^{\top}\cdot x_{t^{*}}^{\left(L+1\right)}.
\end{equation}
where we adopt the unified notation
\begin{equation}
x_{t^{*}}^{\left(L+1\right)}=\begin{cases}
\frac{1}{T}\sum_{t=1}^{T}x_{t}^{\left(L+1\right)} & \qquad\text{average over all the tokens}\\
x_{t^{*}}^{\left(L+1\right)} & \qquad\text{readout from a single token at position \ensuremath{t=t^{*}}}
\end{cases}\label{eq:compact_readout}
\end{equation}
in order to consider different options of token readout, adopted in
different tasks.

\subsection{Posterior distribution}

We are interested in computing the posterior distribution over the
network's weights $\Theta\coloneqq\left(V^{\left(0\right)},\left\{ V^{\left(\ell\right)h}\right\} _{\ell,h=1}^{L,H},a\right)$.
This is analogous to the Gibbs distribution in statistical physics
and is defined as 
\begin{equation}
p\left(\Theta\right)=\frac{1}{Z}\exp\left\{ -\frac{1}{2\mathcal{T}}\sum_{\mu=1}^{P}\left[f\left(x^{\mu},\Theta\right)-y^{\mu}\right]^{2}-\frac{1}{2\sigma^{2}}\left\Vert \Theta\right\Vert ^{2}\right\} \;,\label{eq:posterior_definition}
\end{equation}
where $\left\Vert \cdot\right\Vert $ is the Frobenius norm, and $Z$
indicates the normalization, also called partition function.

\subsection{Attention paths notation\label{subsec:Attention-paths-notation}}

In the main text, we introduce the concept of \emph{attention paths},
by defining a path index
\begin{equation}
\pi\coloneqq\left(h_{1},\ldots,h_{L}\right)\qquad\mathrm{with}\qquad h_{1},\ldots,h_{L}\in\left\{ 1,\ldots,H\right\} \label{eq:path_index}
\end{equation}
which uniquely identifies each possible combination of the head indices,
i.e. each possible path through the attention heads (\Figref{fig:illustration}). The network output
can be written as
\begin{equation}
f=\frac{1}{\sqrt{H^{L}NN_{0}}}\sum_{\pi\in\Pi}V^{\left(\mathrm{eff}\right)\pi}\cdot V^{\left(0\right)}\cdot\xi^{\pi}\label{eq:effective_description-1}
\end{equation}
where $\Pi$ is the set of all possible paths, and we define the \emph{effective
weights} as
\begin{equation}
V^{\left(\mathrm{eff}\right)\pi}\coloneqq\frac{1}{\sqrt{N^{L}}}a\cdot V^{\left(L\right)h_{L}}\cdot V^{\left(L-1\right)h_{L-1}}\cdot\ldots\cdot V^{\left(2\right)h_{2}}\cdot V^{\left(1\right)h_{1}},\quad\quad V^{\left(\mathrm{eff}\right)\pi}\in\mathbb{R}^{1\times N}\label{eq:V_effective-1}
\end{equation}
and the \emph{attentioned input}
\begin{equation}
\xi^{\pi}=\sum_{t_{0},\ldots,t_{L-1}=1}^{T}x_{t_{0}}\Omega_{t_{0}t_{1}}^{\left(1\right)h_{1}}\Omega_{t_{1}t_{2}}^{\left(2\right)h_{2}}\ldots\Omega_{t_{L-2}t_{L-1}}^{\left(L-1\right)h_{L-1}}\Omega_{t_{L-1}t^{*}}^{\left(L\right)h_{L}}\,.\label{eq:attentioned_input-1}
\end{equation}

In order to present our theoretical results and their derivation,
it is useful to also introduce the following notation.

\paragraph{Partial-path indices.}

We indicate with 
\begin{equation}
\pi_{\ell}\coloneqq\left(h_{\ell},h_{\ell+1},\ldots,h_{L}\right)\qquad\mathrm{with}\qquad h_{\ell},\ldots,h_{L}\in\left\{ 1,\ldots,H\right\} \label{eq:compact_head_index-1}
\end{equation}
a collection of head indices, from layer $\ell$ up to layer $L$.
For $\ell=1$, $\pi_{1}\equiv\pi$ is the path index defined above,
Eq.~\ref{eq:path_index}, and in the main text. For $\ell\neq1$, $\pi_{\ell}$
corresponds more generally to a \emph{partial path}, starting from
layer $\ell$. We call the set of such paths $\Pi_{\ell}$. For $\ell=1$,
$\Pi_{1}\equiv\Pi$ is the collection of all complete paths defined
above and in the main text. 

\paragraph{Attentioned input at layer $\ell$.}

We define the \emph{attentioned input at layer} $\ell$, for $\ell=1,\ldots,L$,
as
\begin{equation}
\xi^{\left(\ell\right)\pi_{\ell}}=\xi^{\left(\ell\right)h_{\ell}h_{\ell+1}\ldots h_{L}}\coloneqq\sum_{t_{\ell},t_{\ell+1},\ldots,t_{L}=1}^{T}x_{t_{\ell}}^{\left(\ell\right)}\Omega_{t_{\ell}t_{\ell+1}}^{\left(\ell\right)h_{\ell}}\ldots\Omega_{t_{L},t^{*}}^{\left(L\right)h_{L}},\qquad\xi^{\left(\ell\right)\pi_{\ell}}\in\mathbb{R}^{N}
\end{equation}
were we remind the reader of the compact notation $t^{*}$ defined
in \Eqref{eq:compact_readout}. The attentioned inputs at subsequent
layers are related by
\begin{equation}
\xi^{\left(\ell\right)\pi_{\ell}}=\frac{1}{\sqrt{NH}}\sum_{h_{\ell-1}=1}^{H}V^{\left(\ell-1\right)h_{\ell-1}}\cdot\xi^{\left(\ell-1\right)h_{\ell-1}\pi_{\ell}}
\end{equation}
Recall that $\xi^{\left(\ell-1\right)h_{\ell-1}\pi_{\ell}}\coloneqq\xi^{\left(\ell-1\right)h_{\ell-1}h_{\ell}h_{\ell+1}\ldots h_{L}}$,
according to the notation defined in Eq.~\ref{eq:compact_head_index-1}. 

We also give specific definitions for the cases $\ell=0$ and $\ell=L+1$.
For $\ell=0$, we define $\xi^{\left(0\right)\pi_{1}}\equiv\xi^{\pi}$,
corresponding to the attentioned input defined above, \Eqref{eq:attentioned_input-1},
and in the main text. The following relation holds
\begin{equation}
\xi^{\left(0\right)\pi_{1}}=\frac{1}{\sqrt{N_{0}}}V^{\left(0\right)}\cdot\xi^{\left(1\right)\pi_{1}}\qquad\xi^{\left(0\right)\pi_{1}}\in\mathbb{R}^{N_{0}}\,.
\end{equation}
For $\ell=L+1$ we define
\begin{equation}
\xi^{\left(L+1\right)}=\frac{1}{\sqrt{NH}}\sum_{h_{L}=1}^{H}V^{\left(L\right)h_{L}}\cdot\xi^{\left(L\right)h_{L}}
\end{equation}

\section{Results enunciation\label{apdx_results_enunciation}}

Here we enunciate our theoretical results, for which a derivation
is given in Appendix~\ref{sec:Derivation-of-the} and Appendix~\ref{sec:appendix_oparam_derivation}. 

\subsection{Predictor statistics\label{subsec:Predictor-statistics}}

We derive (Appendix~\ref{sec:Derivation-of-the})  the statistics of the network prediction $f\left(x^{*}\right)\coloneqq f^{*}$
on a new test example $x^{*}$. These are given by
\begin{align}
\left\langle f^{*}\right\rangle  & =k^{\top}\cdot\left(K+\mathcal{T}\mathbb{I}\right)^{-1}\cdot Y\label{eq:mean_pred}\\
\left\langle \left(\delta f^{*}\right)^{2}\right\rangle  & =K_{\mathrm{test}}-k^{\top}\cdot\left(K+\mathcal{T}\mathbb{I}\right)^{-1}\cdot k\label{eq:var_pred}
\end{align}
where $\left\langle \cdot\right\rangle $ denotes the expectation
under the posterior distribution Eq.~\ref{eq:posterior_definition}, and
$\delta f^{*}=f^{*}-\left\langle f^{*}\right\rangle $. The
quantities $K_{\mathrm{test}}\in\mathbb{R}$, $k\in\mathbb{R}^{P}$, and
$K\in\mathbb{R}^{P\times P}$ are defined in terms of a kernel
function $\mathcal{K}:\mathbb{R}^{N_{0}\times T}\times\mathbb{R}^{N_{0}\times T}\to\mathbb{R}$.
For $\mu,\nu=1,\dots,P$ we define $K_{\rm test}\coloneqq\mathcal{K}\left(x^{*},x^{*}\right)$,
$k^{\mu}\coloneqq\mathcal{K}\left(x^{*},x^{\mu}\right)$,
and $K^{\mu\nu}\coloneqq\mathcal{K}\left(x^{\mu},x^{\nu}\right)$.

We derive the following form for the kernel
\begin{equation}
\mathcal{K}\left(x,x'\right)=\frac{1}{H^{L}}\sum_{\pi,\pi'\in\Pi}U^{\pi\pi'}C_{\pi\pi'}\left(x,x'\right)\qquad\mathrm{with}\quad C_{\pi\pi'}\left(x,x'\right)\coloneqq\frac{1}{N_{0}}\xi^{\pi\top}\left(x\right)\cdot\xi^{\pi'}\left(x'\right)\,,\label{eq:kernel_definition}
\end{equation}
where $\xi^{\pi}\left(x\right)$ is the attentioned input corresponding to an input $x$, while
$U$, called \emph{order parameter}, is a path-by-path matrix
of size $\left|\Pi\right|\times\left|\Pi\right|$, where $\left|\Pi\right|=H^{L}$
denotes the size of the set $\Pi$. The value of the order parameter
is determined \emph{self-consistently }by minimizing a scalar function
$S$, called \emph{action} in the physics literature. This is defined as 
\begin{equation}
S\left(\left\{ U^{\left(\ell\right)}\right\} _{\ell=1}^{L+1};X,Y\right)\coloneqq-\mathcal{L}\left(U^{\left(L+1\right)}\right)-\sum_{\ell=1}^{L}\mathcal{L}\left(U^{\left(\ell\right)}\cdot U_{\mathrm{ext}}^{\left(\ell+1\right)-1}\right)+\alpha\text{\ensuremath{\mathcal{E}\left(U^{\left(1\right)};X,Y\right)}}\label{eq:action}
\end{equation}
where, for any matrix $M$, we define 
\begin{equation}
\mathcal{L}\left(M\right)=-\sigma^{-2}\mathrm{Tr}\left(M\right)+\ln\det\left(M\right)
\end{equation}
and 
\begin{equation}
\mathcal{E}\left(U^{\left(1\right)};X,Y\right)=\frac{1}{P}\ln\det\left(K(U^{\left(1\right)};X)+\mathcal{T}\mathbb{I}\right)+\frac{1}{P}Y^{\top}\cdot\left(K(U^{\left(1\right)};X)+\mathcal{T}\mathbb{I}\right)^{-1}\cdot Y\label{eq:apdx_energy_enunciation}
\end{equation}
Note that $K(U^{\left(1\right)};X)$ is simply the kernel defined above, through Eq.~\ref{eq:kernel_definition}, where we have explicitly written its dependence on $U^{\left(1\right)}$ and $X$.
Here we have defined a collection of matrix order parameters $U^{\left(\ell\right)}$
for $\ell=1,\dots,L$, of size $\left|\Pi_{\ell}\right|\times\left|\Pi_{\ell}\right|$,
were $\left|\Pi_{\ell}\right|=H^{L+1-\ell}$ denotes the size of the
set $\Pi_{\ell}$. We called $U^{\left(1\right)}\equiv U$. The order
parameter $U^{\left(L+1\right)}$ is instead a scalar. We also defined
$U_{\mathrm{ext}}^{\left(\ell+1\right)}$ for $\ell=1,\ldots,L$ as
the matrix of size $\left|\Pi_{\ell}\right|\times\left|\Pi_{\ell}\right|$
with elements
\[
U_{\mathrm{ext}}^{\left(\ell+1\right)\pi_{\ell},\pi'_{\ell}}=U_{\mathrm{ext}}^{\left(\ell+1\right)h_{\ell}\pi_{\ell+1},h'_{\ell}\pi'_{\ell+1}} \coloneqq U^{\left(\ell+1\right)\pi_{\ell+1},\pi'_{\ell+1}}\delta_{h_{\ell},h'_{\ell}}
\]
where $\pi_{\ell},\pi'_{\ell}\in\Pi_{\ell}$ and $\pi_{\ell+1},\pi'_{\ell+1}\in\Pi_{\ell+1}$
are the partial-path indices defined in Eq.~\ref{eq:compact_head_index-1},
while $h_{\ell},h_{\ell'}=1,\ldots,H$.

The action must be minimized w.r.t. all of the order parameters. Note
that in the main text we defined $S$ as a function of $U^{\left(1\right)}\equiv U$ alone.
By this we mean $S$ evaluated at its minimum for a fixed $U^{\left(1\right)}$, so
that only the minimization w.r.t. $U^{\left(1\right)}$ is left to perform. 

The self-consistent solution for the order parameters is obtained numerically by minimizing
Eq.~\ref{eq:action} with gradient descent methods. Details on the procedure are given in Appendix~\ref{appendix_order_param_numerics}.

\subsection{Order parameter interpretation\label{subsec:Order-parameter-interpretation}}

We derive (Appendix~\ref{sec:appendix_oparam_derivation}) the following expression for the order parameter in terms
of the network weights
\begin{equation}
U^{\pi\pi'}=\frac{1}{N}\left\langle V^{\left(\mathrm{eff}\right)\pi}\cdot V^{\left(\mathrm{eff}\right)\pi'\top}\right\rangle \,\qquad\qquad\pi,\pi'\in\Pi.
\end{equation}
where $\left\langle \cdot\right\rangle $ denotes statistical averaging
over the posterior distribution Eq.~\ref{eq:posterior_definition}.

\section{Derivation of the predictor statistics\label{sec:Derivation-of-the}}

Here we provide the derivation of our result for the predictor statistics,
reported in Sec.~\ref{subsec:Predictor-statistics}.

\subsection{Partition function\label{subsec:Partition-function}}

Let us recall the posterior distribution (\Eqref{eq:posterior_definition})
\begin{equation}
p\left(\Theta\right)=\frac{1}{Z}\exp\left\{ -\frac{1}{2\mathcal{T}}\sum_{\mu=1}^{P}\left[f\left(x^{\mu},\Theta\right)-y^{\mu}\right]^{2}-\frac{1}{2\sigma^{2}}\left\Vert \Theta\right\Vert ^{2}\right\} \;,\label{eq:posterior_definition-1}
\end{equation}
where $Z$ indicates the normalization, also called partition function.

It is useful to introduce a vector $t\in\mathbb{R}^{P}$ of $P$ auxiliary
variables in order to linearize the squared error in Eq.~\ref{eq:posterior_definition-1}.
The partition function then reads 
\begin{equation}
Z\propto\int\mathcal{D}t\mathcal{D}\Theta\exp\left(\sum_{\mu=1}^{P}\imath t^{\mu}\left(f^{\mu}-y^{\mu}\right)\right)\;,
\end{equation}
where $\imath$ is the imaginary unit, $t^{\mu}$ is the $\mu$-th component of $t$, while $f^{\mu}$
is shorthand for $f\left(x^{\mu},\Theta\right)$, and
we use the symbol ``$\propto$'' to indicate that we are neglecting
multiplicative constants. We also defined the Gaussian integration
measures $\mathcal{D}t=dt\exp\left(-\frac{\mathcal{T}}{2}\left\Vert t\right\Vert ^{2}\right)$
and $\mathcal{D}\Theta=d\Theta\exp\left(-\frac{1}{2\sigma^{2}}\left\Vert \Theta\right\Vert ^{2}\right)$.

Calling $f^{*}\coloneqq f^{P+1}$, we add a \emph{source term} $\propto \imath t^{P+1}f^{P+1}$ to the exponential
in the partition function, corresponding to the test example $x^* \coloneqq x^{P+1}$,
obtaining
\begin{equation}
Z\left[\imath t^{P+1}\right]\propto\int\mathcal{D}t\mathcal{D}\Theta\exp\left(-\imath\sum_{\mu=1}^{P}t^{\mu}y^{\mu}+\sum_{\mu=1}^{P+1}\imath t^{\mu}f^{\mu}\right)\;.\label{eq:partition_function}
\end{equation}
In what follows we write $Z\left[\imath t^{P+1}\right]$ as $Z$, without
writing its argument explicitly. The partition function Eq.~\ref{eq:partition_function}
allows us to obtain the predictor statistics on the test example by
differentiation
\begin{align}
\left\langle f^*\right\rangle \coloneqq \left\langle f^{P+1}\right\rangle  & =\frac{d}{d\left(\imath t^{P+1}\right)}\ln\left(Z\right)\Big|_{t^{P+1}=0}\label{eq:mean_pred_diff}\\
\left\langle \delta f^* \right\rangle \coloneqq \left\langle \delta f^{P+1}\right\rangle  & =\frac{d^{2}}{d\left(\imath t^{P+1}\right)^{2}}\ln\left(Z\right)\Big|_{t^{P+1}=0}\label{eq:var_pred_diff}
\end{align}
where $\delta f^{P+1}=f^{P+1}-\left\langle f^{P+1}\right\rangle $,
and $\left\langle \cdot\right\rangle $ denotes statistical averaging
over the posterior distribution (\Eqref{eq:posterior_definition}). In what follows, we will neglect any multiplicative constants in the partition function, that are independent of $\imath t^{P+1}$, since they are irrelevant for the computation of \Eqref{eq:mean_pred_diff} and \Eqref{eq:var_pred_diff}.

\subsection{Weights integration}

We proceed to compute the partition function (\Eqref{eq:partition_function})
by integrating all of the network weights $\Theta$, and finally the
auxiliary variable $t$. We use the back-propagating kernel renormalization
(BPKR) method \citep{li2021statistical,li2022globally}, which consists in the successive integration
of the weights, starting from the last layer down to the first layer.

Let us write Eq.~\ref{eq:partition_function} as 
\begin{equation}
Z\propto\int\mathcal{D}t\exp\left(-\imath\sum_{\mu=1}^{P}t^{\mu}y^{\mu}\right)I\left(t\right)\;.
\end{equation}
where we define 
\begin{equation}
I\left(t\right)=\int\mathcal{D}\Theta\exp\left(\sum_{\mu=1}^{P+1}\imath t^{\mu}f^{\mu}\right)\label{eq:weights_integral}
\end{equation}
Below we focus on computing $I$$\left(t\right)$. In what follows,
we will write $I\left(t\right)$ as $I$, without writing its argument
explicitly.

\subsubsection{Integration over the readout weights\label{subsec:Integration-over-the-readout}}

We start by integrating over the readout weights $a$. In Eq.~\ref{eq:weights_integral},
we substitute $f^{\mu}=a\cdot\xi^{\left(L+1\right)\mu}$, obtaining
\begin{equation}
I=\int\mathcal{D}\Theta\exp\left(\imath\sum_{\mu=1}^{P+1}a\cdot\xi^{\left(L+1\right)\mu}t^{\mu}\right)
\end{equation}
In what follows, it is useful to define $q^{\left(L+1\right)}=\sum_{\mu=1}^{P+1}t^{\mu}\xi^{\left(L+1\right)\mu}$
and in general 
\begin{equation}
q^{\left(\ell\right)\pi_{\ell}}=\sum_{\mu=1}^{P+1}t^{\mu}\xi^{\left(\ell\right)\pi_{\ell},\mu},\qquad\pi_{\ell}\in\Pi_{\ell}
\end{equation}
We compute the Gaussian integral in the readout weights $a$, obtaining 
\begin{equation}
I\propto\int\mathcal{D}\Theta^{\left(L\right)}\exp\left(-\frac{1}{2}\frac{\sigma^{2}}{N}q^{\left(L+1\right)\top}\cdot q^{\left(L+1\right)}\right)\label{eq:partition_after_readout}
\end{equation}
With $\Theta^{\left(\ell\right)}$ we indicate the collection of weights
$\Theta^{\left(\ell\right)}\coloneqq\left(V^{\left(0\right)},\left\{ V^{\left(\ell'\right)h}\right\} _{\ell',h=1}^{\ell,H}\right)$,
i.e. all weights up to layer $\ell$.

\subsubsection{Integration over the weights at layer $L$\label{subsec:Integration-over-the}}

Plugging $q^{\left(L+1\right)}=\frac{1}{\sqrt{NH}}\sum_{h_{L}=1}^{H}V^{\left(L\right)h_{L}}\cdot q^{\left(L\right)h_{L}}$
into Eq.~\ref{eq:partition_after_readout} we have 
\begin{equation}
I\propto\int\mathcal{D}\Theta^{\left(L\right)}\exp\left(-\frac{1}{2}\frac{\sigma^{2}}{N}\frac{1}{NH}\sum_{h_{L},h_{L}'=1}^{H}q^{\left(L\right)h_{L}\top}V^{\left(L\right)h_{L}\top}\cdot V^{\left(L\right)h'_{L}}q^{\left(L\right)h'_{L}}\right)
\end{equation}
We perform the integral over the value weights $\left\{ V^{\left(L\right)h}\right\} _{h=1}^{H}$,
by noticing that it is given by the product of $N$ identical integrals
of the form
\begin{equation}
\int\left[\prod_{i=1}^{N}\prod_{h_{L}=1}^{H}dV_{1i}^{\left(L\right)h_{L}}\right]\exp\left(-\frac{1}{2\sigma^{2}}\sum_{i,i'=1}^{N}\sum_{h_{L},h'_{L}=1}^{H}V_{1i}^{\left(L\right)h_{L}}\,A_{ih_{L},i'h'_{L}}\,V_{1i'}^{\left(L\right)h'_{L}}\right)
\end{equation}
where we defined
\begin{equation}
A_{ih_{L},i'h'_{L}}\,=\delta_{i,i'}\delta_{h_{L},h'_{L}}+\frac{\sigma^{2}}{N}\frac{\sigma^{2}}{NH}q_{i}^{\left(L\right)h_{L}}q_{i'}^{\left(L\right)h'_{L}}\label{eq:A_matrix_L}
\end{equation}
with $i,i'=1,\ldots,N$ and $h_{L},h'_{L}=1,\dots,H$. Here $q_{i}^{\left(L\right)h_{L}}$
indicates the $i$-th element of the vector $q^{\left(L\right)h_{L}}\in\mathbb{R}^{N}$.
We may consider $A$ as an $NH\times NH$ matrix, whose two indices
run over the pairs $\left(i,h_{L}\right)$ and $\left(i',h'_{L}\right)$.
With this notation, the result of the value weights integration is
\begin{equation}
I\propto\int\mathcal{D}\Theta^{\left(L-1\right)}\exp\left(-\frac{1}{2}N\ln\det\left(A\right)\right)
\end{equation}
Using the matrix determinant lemma\footnote{$\det\left(\mathbb{I}+qq^{\top}\right)=1+q^{\top}\cdot q$, where
$q$ is a vector.}, we find 
\begin{equation}
I\propto\int\mathcal{D}\Theta^{\left(L-1\right)}\exp\left(-\frac{1}{2}N\ln\left(1+R^{\left(L+1\right)}\right)\right)\label{eq:partition_L_before_order_parameter}
\end{equation}
where we defined the scalar $R^{\left(L+1\right)}$
\begin{equation}
R^{\left(L+1\right)}=\frac{\sigma^{2}}{N}\frac{\sigma^{2}}{NH}\sum_{h_{L}=1}^{H}q^{\left(L\right)h_{L}\top}\cdot q^{\left(L\right)h_{L}}\label{eq:order-param_def_L+1}
\end{equation}
We now enforce the identity of \Eqref{eq:order-param_def_L+1} by
Fourier representation of the Dirac delta function, introducing the
auxiliary scalar variable $\text{\ensuremath{U^{\left(L+1\right)}}}$.
We obtain 
\begin{multline}
I\propto\int dR^{\left(L+1\right)}dU^{\left(L+1\right)}\mathcal{D}\Theta^{\left(L-1\right)}\exp\Bigg\{+\frac{1}{2}\frac{N}{\sigma^{2}}\text{\ensuremath{U^{\left(L+1\right)}}}R^{\left(L+1\right)}-\frac{1}{2}N\ln\left(1+R^{\left(L+1\right)}\right)\\
-\frac{1}{2}\frac{\sigma^{2}}{NH}U^{\left(L+1\right)}\sum_{h_{L}=1}^{H}q^{\left(L\right)h_{L}\top}\cdot q^{\left(L\right)h_{L}}\Bigg\}
\end{multline}
In the statistical physics language, $R^{\left(L+1\right)}$ and $U^{\left(L+1\right)}$
are called order parameters. Since $N\to\infty$, we can solve the
integral in $R^{\left(L+1\right)}$ with the saddle point method \citep{fedoryuk1977saddle, fedoryuk1989asymptotic}.
The value of $R^{\left(L+1\right)}$ at the saddle point is
\begin{equation}
R^{\left(L+1\right)}=\sigma^{2}\text{\ensuremath{U^{\left(L+1\right)-1}-1}}
\end{equation}
Therefore, we obtain 
\begin{multline}
I\propto\int dU^{\left(L+1\right)}\mathcal{D}\Theta^{\left(L-1\right)}\exp\Bigg\{\frac{N}{2}\mathcal{L}\left(\text{\ensuremath{U^{\left(L+1\right)}}}\right)\\
-\frac{1}{2}\frac{\sigma^{2}}{NH}\sum_{\pi{}_{L},\pi'_{L}\in\Pi_{L}}U_{\mathrm{ext}}^{\left(L+1\right)\pi_{L},\pi'_{L}}q^{\left(L\right)\pi_{L}\top}\cdot q^{\left(L\right)\pi'_{L}}\Bigg\}\label{eq:partition_L}
\end{multline}
where we defined the ``entropy'' term
\begin{equation}
\mathcal{L}\left(\text{\ensuremath{U^{\left(L+1\right)}}}\right)=-\frac{1}{\sigma^{2}}\text{\ensuremath{U^{\left(L+1\right)}}}+\ln\left(\text{\ensuremath{U^{\left(L+1\right)}}}\right)\label{eq:scalar_entropy}
\end{equation}
an we introduced the $\left|\Pi_{L}\right|\times\left|\Pi_{L}\right|$
matrix $U_{\mathrm{ext}}^{\left(L+1\right)}$, with matrix elements
\begin{equation}
U_{\mathrm{ext}}^{\left(L+1\right)\pi_{L},\pi'_{L}}=U_{\mathrm{ext}}^{\left(L+1\right)h_{L},h'_{L}}=U^{\left(L+1\right)}\delta_{h_{L},h'_{L}}\,.
\end{equation}
Here $\pi_{L},\pi'_{L}\in\Pi_{L}$ are the partial-path indices defined
in Eq.~\ref{eq:compact_head_index-1}, which in this case coincide with
$h_{L},h_{L'}=1,\ldots,H$. While the definition of $U_{\mathrm{ext}}^{\left(L+1\right)}$
may appear superfluous here, it is useful to perform the proof by
induction in Sec.~\ref{subsec:Integration-over-the-2}. 

\subsubsection{Integration over the weights at layer $L-1$\label{subsec:Integration-over-the-1}}

Next, we perform the integration over the weights at layer $L-1$.
The steps are almost identical to those taken in Sec.~\ref{subsec:Integration-over-the},
but will lead to the introduction of a matrix of order parameters.
After this layer, we will be able to provide the results for the integration
of the weights at subsequent layers by induction.

Plugging $q^{\left(L\right)\pi_{L}}=\frac{1}{\sqrt{NH}}\sum_{h_{L-1}=1}^{H}q^{\left(L-1\right)h_{L-1}\pi_{L}}$
into Eq.~\ref{eq:partition_L}, we get
\begin{multline}
I\propto\int dU^{\left(L+1\right)}\mathcal{D}\Theta^{\left(L-1\right)}\exp\Bigg\{+\frac{N}{2}\mathcal{L}\left(\text{\ensuremath{U^{\left(L+1\right)}}}\right)\\
-\frac{\sigma^{2}}{2N^{2}H^{2}}\sum_{h_{L-1},h'_{L-1}=1}^{H}\sum_{\pi_{L},\pi'_{L}\in\Pi_{L}}U_{\mathrm{ext}}^{\left(L+1\right)\pi_{L},\pi'_{L}}\times\\
\times q^{\left(L-1\right)h_{L-1}\pi_{L}\top}\cdot V^{\left(L-1\right)h_{L-1}\top}\cdot V^{\left(L-1\right)\pi'_{L}}\cdot q^{\left(L-1\right)h'_{L-1}\pi'_{L}}\Bigg\}
\end{multline}
Once again we see that the integral in the value weights $\left\{ V^{\left(L-1\right)h}\right\} _{h=1}^{H}$
is given by the product of $N$ identical integrals of the form 
\begin{multline}
\int\left[\prod_{i=1}^{N}\prod_{h_{L-1}=1}^{H}dV_{1i}^{\left(L-1\right)h_{L-1}}\right]\times\\
\times\exp\left(-\frac{1}{2\sigma^{2}}\sum_{i,i'=1}^{N}\sum_{h_{L-1},h'_{L-1}=1}^{H}V_{1i}^{\left(L-1\right)h_{L-1}}\,A_{ih_{L-1},i'h'_{L-1}}\,V_{1i'}^{\left(L-1\right)h'_{L-1}}\right)
\end{multline}
where we defined
\begin{multline}
A_{ih_{L-1},i'h'_{L-1}}\,=\delta_{i,i'}\delta_{h_{L-1},h'_{L-1}}\\
+\frac{\sigma^{2}}{N}\frac{\sigma^{2}}{NH^{2}}\sum_{\pi_{L},\pi'_{L}\in\Pi_{L}}q_{i}^{\left(L-1\right)h_{L-1}\pi_{L}}U_{\mathrm{ext}}^{\left(L+1\right)\pi_{L},\pi'_{L}}q_{i'}^{\left(L-1\right)h'_{L-1}\pi'_{L}}\label{eq:A_matrix_L-1}
\end{multline}
with $i,i'=1,\ldots,N$ and $h_{L},h'_{L}=1,\dots,H$.

Again, we may consider $A$ as an $NH\times NH$ matrix, whose two
indices run over the pairs $\left(i,h_{L-1}\right)$ and $\left(i',h'_{L-1}\right)$.
With this notation, the result of the value weights integration is
\begin{equation}
I\propto\int dU^{\left(L+1\right)}\mathcal{D}\Theta^{\left(L-2\right)}\exp\Bigg(\frac{N}{2}\mathcal{L}\left(\text{\ensuremath{U^{\left(L+1\right)}}}\right)-\frac{1}{2}N\ln\det\left(A\right)\Bigg)
\end{equation}
Using the matrix determinant lemma\footnote{$\det\left(\mathbb{I}+Q\cdot U\cdot Q^{\top}\right)=\det\left(\mathbb{I}+U\cdot Q^{\top}\cdot Q\right)$,
with $Q$ and $U$ being $m\times n$ and $n\times n$ matrices respectively.}, we find 
\begin{equation}
I\propto\int dU^{\left(L+1\right)}\mathcal{D}\Theta^{\left(L-2\right)}\exp\left(\frac{N}{2}\mathcal{L}\left(\text{\ensuremath{U^{\left(L+1\right)}}}\right)-\frac{1}{2}N\ln\det\left(\mathbb{I}+U_{\mathrm{ext}}^{\left(L+1\right)}\cdot R^{\left(L\right)}\right)\right)
\end{equation}
where we introduced the $\left|\Pi_{L}\right|\times\left|\Pi_{L}\right|$
matrix $R^{\left(L\right)}$, with matrix elements defined as
\begin{equation}
R^{\left(L\right)\pi_{L},\pi'_{L}}=\frac{\sigma^{2}}{N}\frac{\sigma^{2}}{NH^{2}}\sum_{h_{L-1}=1}^{H}q^{\left(L-1\right)h_{L-1}\pi_{L}\top}\cdot q^{\left(L-1\right)h_{L-1}\pi'_{L}}\,,\qquad\qquad\pi_{L},\pi'_{L}\in\Pi_{L}\label{eq:order-param_def_L}
\end{equation}
With the same procedure as in Sec.~\ref{subsec:Integration-over-the}, we
introduce the order parameter $R^{\left(L\right)}$ and its conjugate
$U^{\left(L\right)}$, also a $\left|\Pi_{L}\right|\times\left|\Pi_{L}\right|$
matrix. We have
\begin{multline}
I\propto\int dU^{\left(L+1\right)}dR^{\left(L\right)}dU^{\left(L\right)}\mathcal{D}\Theta^{\left(L-2\right)}\exp\Bigg\{\frac{N}{2}\mathcal{L}\left(\text{\ensuremath{U^{\left(L+1\right)}}}\right)\\
+\frac{1}{2}\frac{N}{\sigma^{2}}\ensuremath{\mathrm{Tr}}\left(\text{\ensuremath{U^{\left(L\right)}}}R^{\left(L\right)}\right)-\frac{1}{2}N\ln\det\left(\mathbb{I}+U_{\mathrm{ext}}^{\left(L+1\right)}\cdot R^{\left(L\right)}\right)\\
-\frac{1}{2}\frac{\sigma^{2}}{NH^{2}}\sum_{\pi_{L},\pi'_{L}\in\Pi_{L}}U^{\left(L\right)\pi_{L},\pi'_{L}}\sum_{h_{L-1}=1}^{H}q^{\left(L-1\right)h_{L-1}\pi_{L}\top}\cdot q^{\left(L-1\right)h{}_{L-1}\pi'_{L}}\Bigg\}
\end{multline}
Again, we solve the integral in $R^{\left(L\right)}$ with the saddle
point method. The value of $R^{\left(L\right)}$ at the saddle point
is
\begin{equation}
R^{\left(L\right)}=\sigma^{2}U^{\left(L\right)-1}-U_{\mathrm{ext}}^{\left(L+1\right)-1}
\end{equation}
Therefore we obtain 
\begin{multline}
I\propto\int dU^{\left(L+1\right)}dU^{\left(L\right)}\mathcal{D}\Theta^{\left(L-2\right)}\exp\Bigg\{\frac{N}{2}\mathcal{L}\left(\text{\ensuremath{U^{\left(L+1\right)}}}\right)+\frac{N}{2}\mathcal{L}\left(\text{\ensuremath{U^{\left(L\right)}}\ensuremath{\cdot}}U_{\mathrm{ext}}^{\left(L+1\right)-1}\right)\\
-\frac{1}{2}\frac{\sigma^{2}}{NH^{2}}\sum_{\pi_{L-1},\pi'_{L-1}\in\Pi_{L-1}}U_{\mathrm{ext}}^{\left(L\right)\pi_{L-1},\pi'_{L-1}}q^{\left(L-1\right)\pi_{L-1}\top}\cdot q^{\left(L-1\right)\pi'_{L-1}}\Bigg\}
\end{multline}
where we give a more general definition of the entropy Eq.~\ref{eq:scalar_entropy},
such that it can take a matrix argument
\begin{equation}
\mathcal{L}\left(\text{\ensuremath{U^{\left(L\right)}}\ensuremath{\cdot}}U_{\mathrm{ext}}^{\left(L+1\right)-1}\right)=-\frac{1}{\sigma^2}\mathrm{Tr}\left(\text{\ensuremath{U^{\left(L\right)}}\ensuremath{\cdot}}U_{\mathrm{ext}}^{\left(L+1\right)-1}\right)+\ln\det\left(\text{\ensuremath{U^{\left(L\right)}}\ensuremath{\cdot}}U_{\mathrm{ext}}^{\left(L+1\right)-1}\right)
\end{equation}
and we introduced the $\left|\Pi_{L-1}\right|\times\left|\Pi_{L-1}\right|$
matrix $U_{\mathrm{ext}}^{\left(L\right)}$, with matrix elements
\begin{equation}
U_{\mathrm{ext}}^{\left(L\right)\pi_{L-1},\pi'_{L-1}}=U_{\mathrm{ext}}^{\left(L\right)h_{L-1}\pi_{L},h'_{L-1}\pi'_{L}}=U^{\left(L\right)\pi_{L},\pi'_{L}}\delta_{h_{L-1},h'_{L-1}}\,.
\end{equation}
Here $\pi_{L-1},\pi'_{L-1}\in\Pi_{L-1}$ and $\pi_{L},\pi'_{L}\in\Pi_{L}$
and are the partial-path indices defined in Eq.~\ref{eq:compact_head_index-1},
while $h_{L-1},h'_{L-1}=1,\ldots,H$.

\subsubsection{Integration over the weights at a generic layer $\ell$\label{subsec:Integration-over-the-2}}

We can now compute the integration over the remaining value weights
by induction. We claim that, after integration of the weights at layer
$\ell$, $I$ will have the form 
\begin{multline}
I\propto\int dU^{\left(L+1\right)}dU^{\left(L\right)}\ldots U^{\left(\ell+1\right)}\mathcal{D}\Theta^{\left(\ell-1\right)}\exp\Bigg\{\frac{N}{2}\mathcal{L}\left(\text{\ensuremath{U^{\left(L+1\right)}}}\right)\\
+\frac{N}{2}\mathcal{L}\left(\text{\ensuremath{U^{\left(L\right)}}\ensuremath{\cdot}}U_{\mathrm{ext}}^{\left(L+1\right)-1}\right)+\ldots+\frac{N}{2}\mathcal{L}\left(\text{\ensuremath{U^{\left(\ell+1\right)}}\ensuremath{\cdot}}U_{\mathrm{ext}}^{\left(\ell+2\right)-1}\right)\\
-\frac{1}{2}\frac{\sigma^{2}}{NH^{L+1-\ell}}\sum_{\pi_{\ell},\pi'_{\ell}\in\Pi_{\ell}}U_{\mathrm{ext}}^{\left(\ell+1\right)\pi_{\ell},\pi'_{\ell}}\sum_{h_{\ell}=1}^{H}q^{\left(\ell\right)\pi_{\ell}\top}\cdot q^{\left(\ell\right)\pi'_{\ell}}\Bigg\}\label{eq:partition_induction}
\end{multline}

Here we defined a collection of matrix order parameters, one for each
integrated layer. The order parameter $U^{\left(\ell\right)}$ is
a partial-path-by-partial-path matrix of size $\left|\Pi_{\ell}\right|\times\left|\Pi_{\ell}\right|$,
were $\left|\Pi_{\ell}\right|=H^{L+1-\ell}$ denotes the size of the
set $\Pi_{\ell}$. We also defined $U_{\mathrm{ext}}^{\left(\ell+1\right)}$
as the matrix of size $\left|\Pi_{\ell}\right|\times\left|\Pi_{\ell}\right|$
with elements
\begin{equation}
U_{\mathrm{ext}}^{\left(\ell+1\right)\pi_{\ell},\pi'_{\ell}}=U_{\mathrm{ext}}^{\left(\ell+1\right)h_{\ell}\pi_{\ell+1},h'_{\ell}\pi'_{\ell+1}} \coloneqq U^{\left(\ell+1\right)\pi_{\ell+1},\pi'_{\ell+1}}\delta_{h_{\ell},h'_{\ell}}
\end{equation}
where $\pi_{\ell},\pi'_{\ell}\in\Pi_{\ell}$ and $\pi_{\ell+1},\pi'_{\ell+1}\in\Pi_{\ell+1}$
are the partial-path indices defined in Eq.~\ref{eq:compact_head_index-1},
while $h_{\ell},h_{\ell'}=1,\ldots,H$.

 \Eqref{eq:partition_induction} is verified for layer $\ell=L-1$,
which we derived in Sec.~\ref{subsec:Integration-over-the-1}. The induction
step, integrating over the weights at layer $\ell-1$, is done by
plugging $q^{\left(\ell\right)\pi_{\ell}}=\sum_{h_{\ell-1}=1}^{H}V^{\left(\ell-1\right)h_{\ell-1}}\cdot q^{\left(\ell-1\right)h_{\ell-1}\pi_{\ell}}$
into Eq.~\ref{eq:partition_induction} and applying exactly the same steps
presented in Sec.~\ref{subsec:Integration-over-the-1}.

\subsection{Integration of the auxiliary variable $t$\label{subsec:t_integration}}

After integrating all of the network weights, we have 
\begin{equation}
Z\propto\int\mathcal{D}t\exp\left(-\imath\sum_{\mu=1}^{P}t^{\mu}y^{\mu}\right)I\left(t\right)\;.
\end{equation}
with 
\begin{multline}
I\left(t\right)\propto\int\left[\prod_{\ell=1}^{L+1}dU^{\left(\ell\right)}\right]\exp\Bigg\{\frac{N}{2}\mathcal{L}\left(\text{\ensuremath{U^{\left(L+1\right)}}}\right)+\frac{N}{2}\sum_{\ell=1}^{L}\mathcal{L}\left(\text{\ensuremath{U^{\left(\ell\right)}}\ensuremath{\cdot}}U_{\mathrm{ext}}^{\left(\ell+1\right)-1}\right)\\
-\frac{1}{2}\frac{\sigma^{2}}{NH^{L}}\sum_{\pi_{1},\pi'_{1}\in\Pi_{1}}U^{\left(1\right)\pi_{1},\pi'_{1}}q^{\left(0\right)\pi_{1}\top}\cdot q^{\left(0\right)\pi'_{1}}\Bigg\}\label{eq:partition_induction-1}
\end{multline}
plugging $q^{\left(0\right)\pi_{1}}=\sum_{\mu=1}^{P+1}\xi^{\left(0\right)\pi_{1},\mu}t^{\mu}$
into Eq.~\ref{eq:partition_induction-1}, we see that we need to perform
the following integral in $t\in\mathbb{R}^{P}$
\begin{equation}
\exp\left(-\frac{1}{2}t^{P+1}K_{\mathrm{test}}t^{P+1}\right)\int\left[\prod_{\mu=1}^{P}dt^{\mu}\right]\exp\left(-\frac{1}{2}t^{\top}\cdot\left(K+\mathcal{T}\mathbb{I}\right)\cdot t-\left(t^{P+1}k+\imath Y\right)^{\top}\cdot t\right)\label{eq:integral_t}
\end{equation}
where for convenience we report here the kernel definitions given
in Sec.~\ref{subsec:Predictor-statistics}. The quantities $K_{\mathrm{test}}\in\mathbb{R}$,
$k\in\mathbb{R}^{P}$, and $K\in\mathbb{R}^{P\times P}$ are defined
in terms of a kernel function $\mathcal{K}:\mathbb{R}^{N_{0}\times T}\times\mathbb{R}^{N_{0}\times T}\to\mathbb{R}$.
For $\mu,\nu=1,\dots,P$ we define $K_{\rm test}\coloneqq\mathcal{K}\left(x^*,x^*\right)$,
$k^{\mu}\coloneqq\mathcal{K}\left(x^{*},x^{\mu}\right)$,
and $K^{\mu\nu}\coloneqq\mathcal{K}\left(x^{\mu},x^{\nu}\right)$.
The form of the kernel is
\begin{equation}
\mathcal{K}\left(x,x'\right)=\frac{1}{H^{L}}\sum_{\pi_{1},\pi'_{1}\in\Pi_{1}}U^{\left(1\right)\pi_{1},\pi'_{1}}C_{\pi_{1}\pi'_{1}}\left(x,x'\right)\,,
\end{equation}
with
\begin{equation}
C_{\pi_{1}\pi'_{1}}\left(x,x'\right)\coloneqq\frac{1}{N_{0}}\xi^{\left(0\right)\pi_{1}\top}\left(x\right)\cdot\xi^{\left(0\right)\pi'_{1}}\left(x'\right)\,,
\end{equation}
Computing the Gaussian integral (\Eqref{eq:integral_t}) we obtain 
\begin{multline}
Z\propto\int\left[\prod_{\ell=1}^{L+1}dU^{\left(\ell\right)}\right]\exp\left(-\frac{N}{2}S\left(\left\{ U^{\left(\ell\right)}\right\} _{\ell=1}^{L+1}\right)\right)\\
\exp\left(\frac{1}{2}\imath t^{P+1}\left[K_{\mathrm{test}}-k^{\top}\cdot\left(K+\mathcal{T}\mathbb{I}\right)^{-1}\cdot k\right]\imath t^{P+1}+\imath t^{P+1}k^{T}\cdot\left(K+\mathcal{T}\mathbb{I}\right)^{-1}\cdot Y\right)\label{eq:partition_pre_saddle}
\end{multline}
where we report here for convenience the definition of the action
$S$ given in Sec.~\ref{subsec:Predictor-statistics}
\begin{equation}
S\left(\left\{ U^{\left(\ell\right)}\right\} _{\ell=1}^{L+1}\right)\coloneqq-\mathcal{L}\left(U^{\left(L+1\right)}\right)-\sum_{\ell=1}^{L}\mathcal{L}\left(U^{\left(\ell\right)}\cdot U_{\mathrm{ext}}^{\left(\ell+1\right)-1}\right)+\alpha\text{\ensuremath{\mathcal{E}\left(U^{\left(1\right)}\right)}}\label{eq:action-1}
\end{equation}
with 
\begin{equation}
\mathcal{E}\left(U^{\left(1\right)}\right)=\frac{1}{P}\ln\det\left(K+\mathcal{T}\mathbb{I}\right)+\frac{1}{P}Y^{\top}\cdot\left(K+\mathcal{T}\mathbb{I}\right)^{-1}\cdot Y
\end{equation}

In the limit $N,P\to\infty$, $\frac{P}{N}\to\alpha\in\mathbb{R}^{+}$,
we solve the integrals in $\left\{ U^{\left(\ell\right)}\right\} _{\ell=1}^{L+1}$
with the saddle point method \citep{fedoryuk1977saddle, fedoryuk1989asymptotic}. The partition function
therefore takes the final form 
\begin{equation}
Z\propto\int\exp\left(\frac{1}{2}\imath t^{P+1}\left[K_{\mathrm{test}}-k^{\top}\cdot\left(K+\mathcal{T}\mathbb{I}\right)^{-1}\cdot k\right]\imath t^{P+1}+\imath t^{P+1}k^{\top}\cdot\left(K+\mathcal{T}\mathbb{I}\right)^{-1}\cdot Y\right)\label{eq:action_after_saddle}
\end{equation}
where we recall that $K_{\mathrm{test}}$, $k$, and $K$ all depend
on $U^{\left(1\right)}$, which must be evaluated at the minimum of
the action Eq.~\ref{eq:action-1} with respect to all of its arguments
$\left\{ U^{\left(\ell\right)}\right\} _{\ell=1}^{L+1}$. 

Differentiating by $\imath t^{P+1}$ the partition function Eq.~\ref{eq:action_after_saddle}
(see \Eqref{eq:mean_pred_diff} and \Eqref{eq:var_pred_diff}),
we obtain the results for the predictor mean (\Eqref{eq:mean_pred})
and variance (\Eqref{eq:var_pred}) presented in Sec.~\ref{subsec:Predictor-statistics}.

\section{Derivation of the order parameter interpretation\label{sec:appendix_oparam_derivation}}

Here we provide the derivation of our result on the order parameter
interpretation, exposed in Sec.~\ref{subsec:Order-parameter-interpretation}.
The derivation is almost identical to that for the predictor statistics
given in Sec.~\ref{sec:Derivation-of-the}. What follows below should be
considered as a continuation of Sec.~\ref{sec:Derivation-of-the}, to which
we refer for definitions. 

For convenience, we report here the result we want to derive

\begin{equation}
U^{\left(1\right)\pi_{1}\pi_{1}'}=\frac{1}{N}\left\langle V^{\left(\mathrm{eff}\right)\pi_{1}}\cdot V^{\left(\mathrm{eff}\right)\pi_{1}'\top}\right\rangle \,.
\end{equation}
where $\pi_{1},\pi'_{1}\in\Pi_{1}$ are the path indices defined in
Eq.~\ref{eq:compact_head_index-1}, while $\left\langle \cdot\right\rangle $
denotes statistical averaging over the posterior distribution Eq.~\ref{eq:posterior_definition}.
We also recall the definition of the network effective weights

\begin{equation}
V^{\left(\mathrm{eff}\right)\pi_{1}}\coloneqq\frac{1}{\sqrt{N^{L}}}a\cdot V^{\left(L\right)h_{L}}\cdot V^{\left(L-1\right)h_{L-1}}\cdot\ldots\cdot V^{\left(2\right)h_{2}}\cdot V^{\left(1\right)h_{1}},\quad\quad V^{\left(\mathrm{eff}\right)\pi_{1}}\in\mathbb{R}^{1\times N}
\end{equation}

\subsection{Partition function}

As in Sec.~\ref{subsec:Partition-function}, we start from the partition
function
\begin{equation}
Z\propto\int\mathcal{D}t\mathcal{D}\Theta\exp\left(\sum_{\mu=1}^{P}\imath t^{\mu}\left(f^{\mu}-y^{\mu}\right)\right)\;.
\end{equation}

To the partition function, we add a \emph{source term} $\propto\sum_{\pi_{1}\in\Pi_{1}}V^{\left(\mathrm{eff}\right)\pi_{1}}\cdot q_{*}^{\left(1\right)\pi_{1}}$,
with $q_{*}^{\left(1\right)\pi_{1}}\in\mathbb{R}^{N}$
\begin{equation}
Z\propto\int\mathcal{D}t\mathcal{D}\Theta\exp\left(-\imath\sum_{\mu=1}^{P}t^{\mu}y^{\mu}+\sum_{\mu=1}^{P}\imath t^{\mu}f^{\mu}+\frac{\imath}{\sqrt{NH^{L}}}\sum_{\pi_{1}\in\Pi_{1}}V^{\left(\mathrm{eff}\right)\pi_{1}}\cdot q_{*}^{\left(1\right)\pi_{1}}\right)\;.\label{eq:partition_function-oparam}
\end{equation}
such that differentiating by $q_{*}^{\left(1\right)\pi_{1}}$ allows
us to obtain
\begin{align}
\frac{1}{N}\left\langle V^{\left(\mathrm{eff}\right)\pi}\cdot V^{\left(\mathrm{eff}\right)\pi'\top}\right\rangle  & =-H^{L}\frac{1}{Z}\sum_{i=1}^{N}\frac{dZ}{dq_{*,i}^{\left(0\right)\pi_{1}}dq_{*,i}^{\left(0\right)\pi'_{1}}}\Bigg|_{q_{*}^{\left(0\right)}=0}\label{eq:oparam_diff}
\end{align}
 where $q_{*,i}^{\left(0\right)\pi_{1}}$ indicates the $i$-th component
of the vector $q_{*}^{\left(0\right)\pi_{1}}\in\mathbb{R}^{N}$, while
$q_{*}^{\left(0\right)}\in\mathbb{R}^{N\times H^{L}}$ is the matrix
whose $\pi_{1}$-th component is $q_{*}^{\left(0\right)\pi_{1}}$.

As in Sec.~\ref{sec:Derivation-of-the}, we proceed to compute the partition
function (\Eqref{eq:partition_function-oparam}) by integrating all
of the network weights $\Theta$, and finally the auxiliary variable
$t$. We make the following observation. In Eq.~\ref{eq:partition_function-oparam},
we can write explicitly $f^{\mu}=\frac{1}{\sqrt{NH^{L}}}\sum_{\pi_{1}\in\Pi_{1}}V^{\left(\mathrm{eff}\right)\pi_{1}}\cdot\xi^{\left(1\right)\pi_{1},\mu}$.
Furthermore, as in Sec.~\ref{subsec:Integration-over-the-readout} we can
define $q^{\left(L+1\right)}=\sum_{\mu=1}^{P}t^{\mu}\xi^{\left(L+1\right)\mu}$
and in general 
\begin{equation}
q^{\left(\ell\right)\pi_{\ell}}=\sum_{\mu=1}^{P}t^{\mu}\xi^{\left(\ell\right)\pi_{\ell},\mu},\qquad\pi_{\ell}\in\Pi_{\ell}
\end{equation}
Then Eq.~\ref{eq:partition_function-oparam} takes the form 
\begin{equation}
Z\propto\int\mathcal{D}t\mathcal{D}\Theta\exp\left(-\imath\sum_{\mu=1}^{P}t^{\mu}y^{\mu}+\frac{\imath}{\sqrt{NH^{L}}}\sum_{\pi_{1}\in\Pi_{1}}V^{\left(\mathrm{eff}\right)\pi_{1}}\cdot\left(q^{\left(1\right)\pi_{1}}+q_{*}^{\left(1\right)\pi_{1}}\right)\right)\label{eq:oparam_partition_q}
\end{equation}
Renaming $q^{\left(1\right)\pi_{1}}+q_{*}^{\left(1\right)\pi_{1}}\to q^{\left(1\right)\pi_{1}}$,
we see that the steps for computing Eq.~\ref{eq:oparam_partition_q} are
identical to those in Sec.~\ref{sec:Derivation-of-the}, until the integration
over the input projection weights $V^{\left(0\right)}$.

\subsection{Integration of the input projection weights}

After integrating all of the network weights except the input projection
$V^{\left(0\right)}$ we have 
\begin{equation}
Z\propto\int\mathcal{D}t\exp\left(-\imath\sum_{\mu=1}^{P}t^{\mu}y^{\mu}\right)I\left(t\right)\;.
\end{equation}
with 
\begin{multline}
I\left(t\right)\propto\int\mathcal{D}V{}^{\left(0\right)}\int\left[\prod_{\ell=2}^{L+1}dU^{\left(\ell\right)}\right]\exp\Bigg\{+\frac{N}{2}\mathcal{L}\left(\text{\ensuremath{U^{\left(L+1\right)}}}\right)+\frac{N}{2}\sum_{\ell=2}^{L}\mathcal{L}\left(\text{\ensuremath{U^{\left(\ell\right)}}\ensuremath{\cdot}}U_{\mathrm{ext}}^{\left(\ell+1\right)-1}\right)\\
-\frac{1}{2}\frac{\sigma^{2}}{NH^{L}}\sum_{\pi_{1},\pi'_{1}\in\Pi_{1}}U_{\mathrm{ext}}^{\left(2\right)\pi_{1},\pi'_{1}}q^{\left(1\right)\pi_{1}\top}\cdot q^{\left(1\right)\pi'_{1}}\Bigg\}\label{eq:partition_induction-1-1}
\end{multline}
We now substitute $q^{\left(1\right)\pi_{1}}\to q^{\left(1\right)\pi_{1}}+q_{*}^{\left(1\right)\pi_{1}}$
in Eq.~\ref{eq:partition_induction-1-1}, as well as $q^{\left(1\right)\pi_{1}}=\frac{1}{\sqrt{N_{0}}}\sum V^{\left(0\right)}\cdot q^{\left(0\right)\pi_{1}}$
obtaining

\begin{multline}
I\left(t\right)\propto\int\mathcal{D}V{}^{\left(0\right)}\int\left[\prod_{\ell=2}^{L+1}dU^{\left(\ell\right)}\right]\exp\Bigg\{+\frac{N}{2}\mathcal{L}\left(\text{\ensuremath{U^{\left(L+1\right)}}}\right)+\frac{N}{2}\sum_{\ell=2}^{L}\mathcal{L}\left(\text{\ensuremath{U^{\left(\ell\right)}}\ensuremath{\cdot}}U_{\mathrm{ext}}^{\left(\ell+1\right)-1}\right)\\
-\frac{1}{2}\frac{\sigma^{2}}{NH^{L}}\sum_{\pi_{1},\pi'_{1}\in\Pi_{1}}U_{\mathrm{ext}}^{\left(2\right)\pi_{1},\pi'_{1}}q_{*}^{\left(1\right)\pi_{1}\top}\cdot q_{*}^{\left(1\right)\pi'_{1}}\\
-\frac{1}{2}\frac{\sigma^{2}}{NN_{0}H^{L}}\sum_{\pi_{1},\pi'_{1}\in\Pi_{1}}U_{\mathrm{ext}}^{\left(2\right)\pi_{1},\pi'_{1}}q^{\left(0\right)\pi_{1}\top}\cdot V^{\left(0\right)\top}\cdot V^{\left(0\right)}\cdot q^{\left(0\right)\pi'_{1}}\\
-\frac{\sigma^{2}}{N\sqrt{N_{0}}H^{L}}\sum_{\pi_{1},\pi'_{1}\in\Pi_{1}}U_{\mathrm{ext}}^{\left(2\right)\pi_{1},\pi'_{1}}q_{*}^{\left(1\right)\pi_{1}\top}\cdot V^{\left(0\right)}\cdot q^{\left(0\right)\pi'_{1}}\Bigg\}\label{eq:partition_induction-1-1-1}
\end{multline}
The integral in $V^{\left(0\right)}$ in Eq.~\ref{eq:partition_induction-1-1-1}
has the form

\begin{multline}
\int\left[\prod_{i=1}^{N}\prod_{j=1}^{N_{0}}dV_{ij}^{\left(0\right)}\right]\exp\left(-\frac{1}{2\sigma^{2}}\sum_{i=1}^{N}\sum_{j,j'=1}^{N_{0}}V_{ij}^{\left(0\right)}\,A_{j,j'}\,V_{ij'}^{\left(0\right)}-\sum_{i=1}^{N}\sum_{j=1}^{N_{0}}J_{ij}V_{ij}^{\left(0\right)}\right)\label{eq:V0_integral}
\end{multline}
where we defined
\begin{gather}
A_{j,j'}=\delta_{j,j'}+\frac{\sigma^{2}}{N}\frac{\sigma^{2}}{N_{0}H^{L}}\sum_{\pi_{1},\pi'_{1}\in\Pi_{1}}U_{\mathrm{ext}}^{\left(2\right)\pi_{1},\pi'_{1}}q_{j}^{\left(0\right)\pi_{1}}q_{j'}^{\left(0\right)\pi'_{1}}\\
J_{ij}=\frac{\sigma^{2}}{N\sqrt{N_{0}}H^{L}}\sum_{\pi_{1},\pi'_{1}\in\Pi_{1}}U_{\mathrm{ext}}^{\left(2\right)\pi_{1},\pi'_{1}}q_{*,i}^{\left(1\right)\pi_{1}}q_{j}^{\left(0\right)\pi'_{1}}
\end{gather}
where $i=1,\ldots,N$, while $j,j'=1,\ldots,N_{0}$, and with the
notation $q_{*,i}^{\left(1\right)\pi_{1}}$ and $q_{i}^{\left(0\right)\pi{}_{1}}$
we indicate the $i$-th component of the vectors $q_{*}^{\left(1\right)\pi_{1}}$
and $q^{\left(0\right)\pi_{1}}$ respectively. 

We may consider $A_{j,j'}$ as the elements of an $N_{0}\times N_{0}$
matrix $A$. With this notation, the partition function after performing
the integral Eq.~\ref{eq:V0_integral} is
\begin{multline}
Z\propto\int\mathcal{D}t\exp\left(-\imath\sum_{\mu=1}^{P}t^{\mu}y^{\mu}\right)\int\left[\prod_{\ell=2}^{L+1}dU^{\left(\ell\right)}\right]\times\\
\times\exp\Bigg\{\frac{N}{2}\mathcal{L}\left(\text{\ensuremath{U^{\left(L+1\right)}}}\right)+\frac{N}{2}\sum_{\ell=2}^{L}\mathcal{L}\left(\text{\ensuremath{U^{\left(\ell\right)}}\ensuremath{\cdot}}U_{\mathrm{ext}}^{\left(\ell+1\right)-1}\right)-\frac{N}{2}\ln\det\left(A\right)\\
-\frac{1}{2}\frac{\sigma^{2}}{NH^{L}}\sum_{\pi_{1},\pi'_{1}\in\Pi_{1}}q_{*}^{\left(1\right)\pi'_{1}\top}q_{*}^{\left(1\right)\pi_{1}}\left(U_{\mathrm{ext}}^{\left(2\right)\pi_{1},\pi'_{1}}-\frac{\sigma^{2}}{N}\frac{\sigma^{2}}{N_{0}H^{L}}\sum_{\rho_{1},\rho_{1}'\in\Pi_{1}}U_{\mathrm{ext}}^{\left(2\right)\pi_{1},\rho_{1}}q_{j}^{\left(0\right)\rho_{1}}A_{j,j'}^{-1}q_{j'}^{\left(0\right)\rho'_{1}}U_{\mathrm{ext}}^{\left(2\right)\rho'_{1},\pi'_{1}}\right)\Bigg\}\label{eq:partition_induction-1-1-1-1}
\end{multline}

We now differentiate the partition function in \Eqref{eq:partition_induction-1-1-1-1}
by $q_{*}^{\left(1\right)}$, as specified by \Eqref{eq:oparam_diff},
obtaining 
\begin{multline}
\frac{1}{N}\left\langle V^{\left(\mathrm{eff}\right)\pi_{1}}\cdot V^{\left(\mathrm{eff}\right)\pi'_{1}\top}\right\rangle =\frac{1}{Z\left(q_{*}^{\left(0\right)}=0\right)}\int\mathcal{D}t\exp\left(-\imath\sum_{\mu=1}^{P}t^{\mu}y^{\mu}\right)\int\left[\prod_{\ell=2}^{L+1}dU^{\left(\ell\right)}\right]\times\\
\sigma^{2}\left(U_{\mathrm{ext}}^{\left(2\right)\pi_{1},\pi'_{1}}-\frac{\sigma^{2}}{N}\frac{\sigma^{2}}{N_{0}H^{L}}\sum_{\rho_{1},\rho_{1}'\in\Pi_{1}}U_{\mathrm{ext}}^{\left(2\right)\pi_{1},\rho_{1}}q_{j}^{\left(0\right)\rho_{1}}A_{j,j'}^{-1}q_{j'}^{\left(0\right)\rho'_{1}}U_{\mathrm{ext}}^{\left(2\right)\rho'_{1},\pi'_{1}}\right)\times\\
\times\exp\left(+\frac{N}{2}\mathcal{L}\left(\text{\ensuremath{U^{\left(L+1\right)}}}\right)+\frac{N}{2}\sum_{\ell=2}^{L}\mathcal{L}\left(\text{\ensuremath{U^{\left(\ell\right)}}\ensuremath{\cdot}}U_{\mathrm{ext}}^{\left(\ell+1\right)-1}\right)-\frac{N}{2}\ln\det\left(A\right)\right)
\end{multline}
The remaining steps are the same as in Sec.~\ref{sec:Derivation-of-the}.
We use the matrix determinant lemma\footnote{$\det\left(\mathbb{I}+Q\cdot U\cdot Q^{\top}\right)=\det\left(\mathbb{I}+U\cdot Q^{\top}\cdot Q\right)$,
with $Q$ and $U$ being $m\times n$ and $n\times n$ matrices respectively.} and the Woodbury matrix identity\footnote{$\left(\mathbb{I}+Q\cdot U\cdot Q^{\top}\right)^{-1}=\mathbb{I}-Q\cdot\left(\mathbb{I}+U\cdot Q^{\top}\cdot Q\right)^{-1}\cdot U\cdot Q^{T}$,
with $Q$ and $U$ being $m\times n$ and $n\times n$ matrices respectively.} respectively to express $\ln\det\left(A\right)$ and $A^{-1}$ in
terms of a $\left|\Pi_{1}\right|\times\left|\Pi_{1}\right|$ matrix
$R^{\left(1\right)}$ with elements
\[
R^{\left(1\right)\pi_{1},\pi'_{1}}=\frac{\sigma^{2}}{N}\frac{\sigma^{2}}{N_{0}H^{L}}q^{\left(0\right)\pi_{1}\top}\cdot q^{\left(0\right)\pi'_{1}}\qquad\pi_{1},\pi'_{1}\in\Pi_{L}
\]
whose identity we enforce by Fourier representation of the Dirac delta
function, introducing the auxiliary $\left|\Pi_{1}\right|\times\left|\Pi_{1}\right|$
matrix $\text{\ensuremath{U^{\left(1\right)}}}$. The result of these
operations is
\begin{multline}
\frac{1}{N}\left\langle V^{\left(\mathrm{eff}\right)\pi_{1}}\cdot V^{\left(\mathrm{eff}\right)\pi_{1}'\top}\right\rangle =\frac{1}{Z\left(q_{*}^{\left(0\right)}=0\right)}\int\mathcal{D}t\exp\left(-\imath\sum_{\mu=1}^{P}t^{\mu}y^{\mu}\right)\int\left[\prod_{\ell=1}^{L+1}dU^{\left(\ell\right)}\right]dR^{\left(1\right)}\times\\
\sigma^{2}\left[U_{\mathrm{ext}}^{\left(2\right)}\left(\mathbb{I}-R^{\left(1\right)}\cdot U_{\mathrm{ext}}^{\left(2\right)}+R^{\left(1\right)}\cdot\left(\mathbb{I}+U_{\mathrm{ext}}^{\left(2\right)}\cdot R^{\left(1\right)}\right)^{-1}\cdot U_{\mathrm{ext}}^{\left(2\right)}\cdot R^{\left(1\right)}\cdot U_{\mathrm{ext}}^{\left(2\right)}\right)\right]^{\pi_{1},\pi'_{1}}\times\\
\times\exp\Bigg\{+\frac{N}{2}\mathcal{L}\left(\text{\ensuremath{U^{\left(L+1\right)}}}\right)+\frac{N}{2}\sum_{\ell=2}^{L}\mathcal{L}\left(\text{\ensuremath{U^{\left(\ell\right)}}\ensuremath{\cdot}}U_{\mathrm{ext}}^{\left(\ell+1\right)-1}\right)\\
+\frac{1}{2}\frac{N}{\sigma^{2}}\ensuremath{\mathrm{Tr}}\left(\text{\ensuremath{U^{\left(1\right)}}}R^{\left(1\right)}\right)-\frac{1}{2}N\ln\det\left(\mathbb{I}+U_{\mathrm{ext}}^{\left(2\right)}\cdot R^{\left(1\right)}\right)\\
-\frac{1}{2}\frac{\sigma^{2}}{NH^{L}}\sum_{\pi_{1},\pi'_{1}\in\Pi_{1}}U^{\left(1\right)\pi_{1},\pi'_{1}}q^{\left(0\right)\pi_{1}\top}\cdot q^{\left(0\right)\pi'_{1}}\Bigg\}\label{eq:o_param_before_R1}
\end{multline}
As in Sec.~\ref{sec:Derivation-of-the}, we solve the integral in $R^{\left(1\right)}$
with the saddle point method. The value of $R^{\left(1\right)}$ at
the saddle point is
\begin{equation}
R^{\left(1\right)}=\sigma^{2}U^{\left(1\right)-1}-U_{\mathrm{ext}}^{\left(2\right)-1}
\end{equation}
Plugging this back into Eq.~\ref{eq:o_param_before_R1} we obtain

\begin{multline}
\frac{1}{N}\left\langle V^{\left(\mathrm{eff}\right)\pi_{1}}\cdot V^{\left(\mathrm{eff}\right)\pi_{1}'\top}\right\rangle =\frac{1}{Z\left(q_{*}^{\left(0\right)}=0\right)}\int\mathcal{D}t\exp\left(-\imath\sum_{\mu=1}^{P}t^{\mu}y^{\mu}\right)\int\left[\prod_{\ell=1}^{L+1}dU^{\left(\ell\right)}\right]\times\\
\times U^{\left(1\right)\pi_{1},\pi'_{1}}\exp\Bigg\{\frac{N}{2}\mathcal{L}\left(\text{\ensuremath{U^{\left(L+1\right)}}}\right)+\frac{N}{2}\sum_{\ell=1}^{L}\mathcal{L}\left(\text{\ensuremath{U^{\left(\ell\right)}}\ensuremath{\cdot}}U_{\mathrm{ext}}^{\left(\ell+1\right)-1}\right)\\
-\frac{1}{2}\frac{\sigma^{2}}{NH^{L}}\sum_{\pi_{1},\pi'_{1}\in\Pi_{1}}U^{\left(1\right)\pi_{1},\pi'_{1}}q^{\left(0\right)\pi_{1}\top}\cdot q^{\left(0\right)\pi'_{1}}\Bigg\}\label{eq:o_param_before_R1-1}
\end{multline}

\subsection{Integration of the auxiliary variable $t$}

The calculation of the integral in $t$ follows that in Sec.~\ref{subsec:t_integration}.
We obtain
\begin{equation}
\frac{1}{N}\left\langle V^{\left(\mathrm{eff}\right)\pi_{1}}\cdot V^{\left(\mathrm{eff}\right)\pi_{1}'\top}\right\rangle =\left\langle U^{\left(1\right)\pi_{1},\pi'_{1}}\right\rangle _{U^{\left(1\right)}\sim p\left(U^{\left(1\right)}\right)}\label{eq:o_param_before_R1-1-1}
\end{equation}
where $\left\langle \cdot\right\rangle _{U^{\left(1\right)}\sim p\left(U^{\left(1\right)}\right)}$
denotes the expectation under the distribution
\begin{equation}
p\left(U^{\left(1\right)}\right)\propto\int\left[\prod_{\ell=2}^{L+1}dU^{\left(\ell\right)}\right]\exp\left(-\frac{N}{2}S\left(\left\{ U^{\left(\ell\right)}\right\} _{\ell=1}^{L+1}\right)\right)
\end{equation}
where the action $S$ is the same defined in Sec.~\ref{subsec:t_integration},
Eq.~\ref{eq:action-1}. Exactly as in Sec.~\ref{subsec:t_integration}, we compute
the expectation using the saddle point method, under the limit $N,P\to\infty$,
$\frac{P}{N}\to\alpha\in\mathbb{R}^{+}$. We therefore arrive at the
final result 
\begin{equation}
\frac{1}{N}\left\langle V^{\left(\mathrm{eff}\right)\pi_{1}}\cdot V^{\left(\mathrm{eff}\right)\pi_{1}'\top}\right\rangle =U^{\left(1\right)\pi_{1},\pi'_{1}}
\end{equation}
where $U^{\left(1\right)}$ is the value at the saddle point of $S$.

\newpage
\part{Experiments\label{apdx_experiments}}

\section{Numerical evaluation of the order parameter\label{appendix_order_param_numerics}}

The order parameter in our theory is defined self-consistently as
the minimum of an action, Eq.~\ref{eq:action}. We determine this order parameter numerically,
by minimizing the action using the Adam optimizer \citep{KingmaB14}.
As we describe in Appendix \ref{apdx_optimal_temperature}, for each $N$ (out of 10 choices among $N \in \{a\,10^{b}, 10^4; 
a\in\left\{ 1,2,5\right\}, b\in\left\{1,2,3\right\}\}$), we search for the optimal temperature among 10 choices $\mathcal{T} \in \{a\,10^{-b}, 1.0, 1.5 ; a\in\left\{1,2.5,5,7.5\right\}, b\in\left\{1,2\right\}\}$,
resulting in 100 total configurations.
Each such run takes less than 12 hours on a single A100-40GB GPU.
The learning rate is optimized for each configuration by sweeping among $ \{10^{-4}, a\,10^{-b}; a\in\left\{1,5,8\right\}, b\in\left\{0,1,2,3\right\}\}$ for 10 iterations at the beginning of each run, and by selecting the one that achieves the lowest energy term (averaged over these 10 first optimization steps).

\section{Hamiltonian Monte Carlo sampling\label{appendix_Hamiltonian_Montecarlo}}

We sample the network weights from the posterior distribution (Eq.~\ref{eq:posterior_definition})
using Hamiltonian Monte Carlo sampling \citep{betancourt2018conceptual}. Specifically,
we use the NumPyro implementation of the No U-Turn Sampler (NUTS) \citep{phan2019composable,bingham2019pyro}.

For $N\leq100$, we run $10$ independent chains, consisting of $1000$
warm-up steps and $1000$ sampling steps. We keep one sample every
$10$, for a total of $1000$ samples. Due to limited computational
resources, the number of samples is smaller and varies for $N>100$,
while we did not take samples at all for very large $N\geq 1000$. Note
however, that the large $N$ regime is the least relevant to sample,
since the network is approaching the GP limit. The most important validation
of our theory is performed for the smaller values of $N$.

Each run sampling a model at a given $N$ takes less than 12 hours on a single A100-40GB GPU. The number of runs to sample all model widths for  the hidden Markov chain classification task and the one-shot image classification task is 12.

Regarding the temperature $\mathcal{T}$ of the Bayesian posterior \Eqref{eq:posterior_definition}, this is set differently depending on the task. For the HMC task, we find the temperature $\mathcal{T}$ to
not be relevant for improving the network's classification accuracy. We therefore
set it to a small, but finite value of $\mathcal{T}=0.01$. In contrast, for the one-shot image classification task,
we find tuning the temperature to be particularly important to prevent overfitting. Therefore, we always tune the temperature to the value giving the optimal classification accuracy for the given network depth $N$. We refer to Appendix~\ref{apdx_optimal_temperature} for details on the temperature values and its optimization process.

\section{Hidden Markov chain classification task}

\begin{figure}
    \centering
    \hspace{-1em}\includegraphics[width=1.0\textwidth]{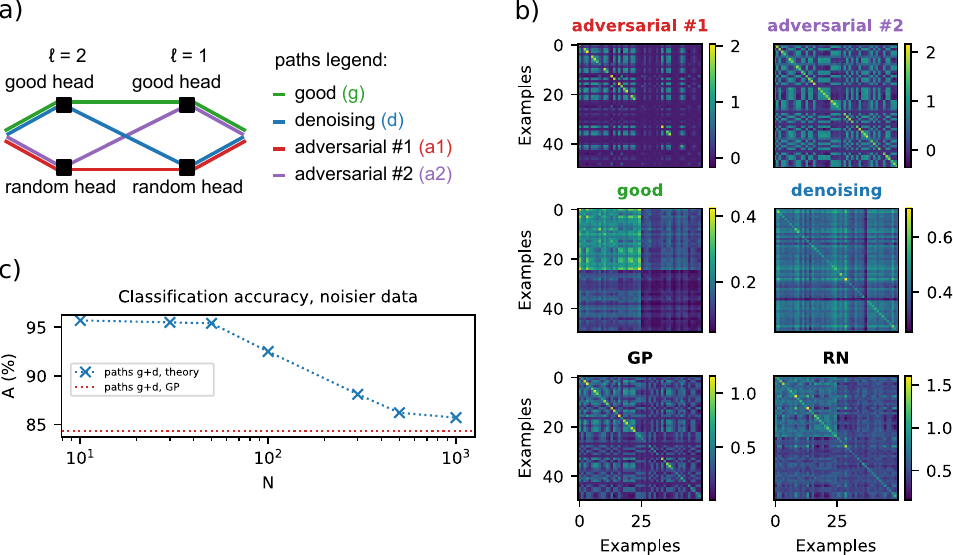}
    \caption{Hidden Markov chain task. \textbf{(a)} Schematics of the network and
    its attention paths. \textbf{(b)} Kernels. Same-path kernels associated with the 4 paths shown in (a), total kernel in the GP limit, and total kernel for $N=10$ in the renormalized regime (RN). Examples on both the $x$ and $y$ axes are ordered by class (the first half correspond to the first class, the second half correspond to the second class). \textbf{(c)} Classification accuracy for a network consisting of only the good and denoising paths, $\sigma_{\parallel}=1$ and $\sigma_{\perp}=5$. The figure is analogous to \Figref{fig:markov}(f), with the difference that we have replaced the random head involved in the denoising path with a uniform attention head.}
    \label{fig:markov_appendix}
\end{figure}

Here we give additional details on the hidden Markov chain (HMC) classification
task.

\subsection{Task definition\label{sec:markov_def_appendix}}

Here we recall the task definition, providing a few additional details. 

The $\mu$-th example in the dataset corresponds to an hidden Markov
chain $q_{1}^{\mu},\ldots,q_{T}^{\mu}$ of length $T=30$, alternating
between two hidden states, $q_{t}^{\mu}\in\left\{ +,-\right\} $.
The transition probability to the opposite state ($"\pm"\to"\mp"$)
is $p^{\mu}$. In other words, describing the ``$\pm$'' states
as one hot vectors $+:\left(\begin{array}{c}
1\\
0
\end{array}\right)$ and $-:\left(\begin{array}{c}
0\\
1
\end{array}\right)$, the transition probability matrix of the hidden Matkov chain is
\[
\left(\begin{array}{cc}
1-p^{\mu} & p^{\mu}\\
p^{\mu} & 1-p^{\mu}
\end{array}\right)\,.
\]
The $\mu$-th chain can belong to one of two classes, labeled $y^{\mu}=\pm1$,
depending on whether $p^{\mu}=0.3$ or $p^{\mu}=0.7$ respectively.

The network is presented with visible states - the input tokens -
which are a noisy, higher dimensional representation of the hidden
states. These are given by 
\[
x_{t}^{\mu}=v_{q_{t}^{\mu}}+\eta_{t}^{\mu}
\]
Here $v_{\pm}\in\mathbb{R}^{N_{0}}$ are two orthogonal feature vectors
corresponding to the states ``$\pm$''. We set $N_{0}=200$ and 
\[
v_{+,i}=\begin{cases}
\sqrt{2} & \text{if \ensuremath{i\leq100}}\\
0 & \text{otherwise}
\end{cases}\qquad\qquad v_{+,i}=\begin{cases}
0 & \text{if \ensuremath{i\leq100}}\\
\sqrt{2} & \text{otherwise}
\end{cases}
\]
The term $\eta_{t}^{\mu}$ is a zero-mean Gaussian noise, with $\langle\eta_{t}^{\mu}\eta_{t'}^{\mu'T}\rangle=\delta_{\mu,\mu'}\delta_{t,t'}(\sigma_{\parallel}^{2}P_{\parallel}^{\top}\cdot P_{\parallel}+\sigma_{\perp}^{2}P_{\perp}^{\top}\cdot P_{\perp})$,
where $P_{\parallel}$ and $P_{\perp}$ are the projectors along the
subspace parallel or perpendicular to the plane spanned by $v_{+}$
and $v_{-}$.

We also preprocess the dataset in two ways: adding to each chain a
beginning-of-sentence (bos) token of zeros at position $t=0$, and
concatenating to each token a one-hot positional encoding vector of
size $T+1$ (i.e. the number of tokens including the bos token).

We use $P=100$ examples for training and $P^{*}=1000$ examples for
testing. For this task, we find the temperature $\mathcal{T}$ to
not be relevant for improving the network performance. We therefore
set it to a small, but finite value of $\mathcal{T}=0.01$. A finite
value of $\mathcal{T}$ is required to sample the predictor statistics,
hence comparing the theoretical results with samples.

\subsection{Query and Key weights initialization\label{apdx_query-key-init}}

Here we give the details of the initialization of the fixed query
and key weights.

We recall that we consider a network of $L=2$ layers and $H=2$ heads
per layer, with readout from the first token (i.e., $t^{*}=1$). The
network has a total of $4$ attention paths (\Figref{fig:markov_appendix}(a)). For the first head of
each layer, we make a good choice of the fixed query and key weights,
which defines a ``good'' attention path, achieving a good classification
accuracy (cf. Sec.~\ref{subsec:Hidden-Markov-chain} in the main text). The remaining heads are initialized
at random. Below we give the initialization details of these good
and random heads.

Let us recall here the definition of the attention matrix $\Omega^{\left(\ell\right)h}\in\mathbb{R}^{T\times T}$
for head $h$ at layer $\ell$. Its matrix elements are defined as
\[
\Omega_{st}^{\left(\ell\right)h}=\zeta\left(\frac{1}{N_{0}\sqrt{G}}x_{s}^{\top}\cdot W_{K}^{\left(\ell\right)h\top}\cdot W_{Q}^{\left(\ell\right)h}\cdot x_{t}\right),\qquad\qquad W_{Q}^{\left(\ell\right)h},W_{K}^{\left(\ell\right)h}\in\mathbb{R}^{G\times\left(N_{0}+T+1\right)}
\]
where $s,t=0,\dots,T$, while $\zeta$ is the softmax function, applied
along the direction of the token index $s$, and $G$ is the dimension
of the query-key feature space. We directly initialize the query-key
matrix product $W^{\left(\ell\right)h}\coloneqq W_{K}^{\left(\ell\right)h\top}\cdot W_{Q}^{\left(\ell\right)h}$.
Note that $W^{\left(\ell\right)h}$ is an $\left[N_{0}+\left(T+1\right)\right]\times\left[N_{0}+\left(T+1\right)\right]$
matrix, because we have appended a one-hot positional encoding vector
to the $N_{0}$-dimensional input tokens (see Sec.~\ref{sec:markov_def_appendix}).
In order to define the heads, it is convenient to decompose $W^{\left(\ell\right)h}$
into the block structure
\[
W^{\left(\ell\right)h}=\beta\left(\begin{array}{cc}
W_{\text{ff}}^{\left(\ell\right)h} & W_{\text{fp}}^{\left(\ell\right)h}\\
W_{\text{pf}}^{\left(\ell\right)h} & W_{\text{pp}}^{\left(\ell\right)h}
\end{array}\right)
\]
where $W_{\text{pp}}^{\left(\ell\right)h}\in\mathbb{R}^{\left(T+1\right)\times\left(T+1\right)}$
acts only on the one-hot positional encoding subspace, $W_{\text{ff}}^{\left(\ell\right)h}\in\mathbb{R}^{N_{0}\times N_{0}}$
acts only on the subspace of the tokens' ``features'', and $W_{\text{fp}}^{\left(\ell\right)h},W_{\text{pf}}^{\left(\ell\right)h\top}\in\mathbb{R}^{N_{0}\times\left(T+1\right)}$
mix these two subspaces. The scalar $0<\beta<\infty$ is a parameter
controlling the ``hardness'' of the softmax function (for $\beta\to\infty$,
the softmax becomes a hardmax). We set it to $\beta=10$ for all heads. 

\subsubsection{Good heads}

Let us define the good heads, i.e. $W^{\left(\ell\right)h}$ for $h=1$
and $\ell=1,2$. Note that the goal here is not to define heads that
are provably good at solving the task, but rather to make a good guess
for their initialization, based on our knowledge of the nature nature of the task.

\paragraph{First layer.}

For the head $h=1$, $\ell=1$ we define
\[
W_{\text{ff}}^{\left(1\right)1}=\frac{1}{N_{0}}\left(v^{+}-v^{-}\right)\left(v^{+}-v^{-}\right)^{\top},\qquad\qquad W_{\text{fp}}^{\left(1\right)1}=W_{\text{pf}}^{\left(1\right)1}=\boldsymbol{0}
\]
and
\[
\left[W_{\text{pp}}^{\left(1\right)1}\right]_{tt'}=\text{\ensuremath{\frac{3}{2}}}\delta_{0,t}+1\delta_{t,t^{'}+1}\qquad\qquad t,t'=1,\ldots,T
\]
where $\left[W_{pp}^{\left(1\right)1}\right]_{tt'}$ is the component
of $W_{pp}^{\left(1\right)1}$ at indices $t$,$t'$. 

\paragraph{Second layer. }

The head $h=1$, $\ell=2$ implements uniform attention $\Omega_{st}^{\left(2\right)1}=\frac{1}{T+1}$,
$\forall s,t=0,\ldots,T$. It is defined by $W_{\text{ff}}^{\left(1\right)1}=W_{\text{fp}}^{\left(1\right)1}=W_{\text{pf}}^{\left(1\right)1}=\boldsymbol{0}$,
and $W_{\text{pp}}^{\left(1\right)1}=\boldsymbol{1}$.

As discussed in the main text, the attention path defined by the good
heads achieves a good classification accuracy. We can give an intuition
as to why this is the case. Intuitively speaking, in the limit of
a hardmax attention $\left(\beta\to\infty\right)$ and no noise $\left(\sigma_{\parallel},\sigma_{\perp}\to0\right)$,
the attention path is ``counting'' the number of times a token has
remained in the same state after a new step in the Markov chain. Indeed,
the first head ``detects'' when there has not been a change of state
between adjacent tokens. It does so by either attending nearby tokens
if and only if they are in the same state, or attending ``nothing''
(in the sense of the zero beginning-of-sentence token). Then, the
second head sums over the tokens attended by the first head,
thereby ``counting'' the number of times a token has not changed
state. More generally, outside the above mentioned limit, we can say
that the good attention path focuses on the two most relevant pieces
of information needed to solve the task: First, it preferentially
pays attention to nearby tokens, which is important because of the
memoryless nature of the Markov process; Second, it is able to detect
the type of transition occurrying between nearby tokens (i.e. remaining in the same state,
or changing state), which is important to distinguish between the
two classes, since they differ by their transition probability.

\subsubsection{Random heads}

Let us define the random heads, i.e. $W^{\left(\ell\right)h}$ for
$h=2$ and $\ell=1,2$. These are initialized with Gaussian
identically and independently distributed entries
\begin{flalign*}
\left[W_{\text{pp}}^{\left(\ell\right)h}\right]_{i,j}\sim\frac{1}{N_{0}}\mathcal{N}\left(0,1\right)\quad & \qquad\left[W_{\text{pp}}^{\left(\ell\right)h}\right]_{t,t'}\sim\mathcal{N}\left(0,1\right)\\
\left[W_{\text{pf}}^{\left(\ell\right)h}\right]_{t,j}\sim\frac{1}{\sqrt{N_{0}}}\mathcal{N}\left(0,1\right)\, & \qquad\left[W_{\text{fp}}^{\left(\ell\right)h}\right]_{i,t'}\sim\frac{1}{\sqrt{N_{0}}}\mathcal{N}\left(0,1\right)
\end{flalign*}
 $\forall i,j=1,\ldots N_{0}$ and $\forall t,t'=0,\ldots,T$. Note
that we take care of proper normalization of the above matrices, depending
on which subspaces they act upon (i.e. the ``features'' or the ``one-hot
positions'' subspaces).

As mentioned in the main text (Sec.~\ref{subsec:Hidden-Markov-chain}), the random heads introduce
three additional paths: two adversarial paths, deteriorating the performance
of the good path, and one ``denoising'' path, improving the good
path performance. We can get an intuition of why this is so by looking
at their associated same-path kernels, \Figref{fig:markov_appendix}(b). We can see that
both the good-path and the two adversarial-path kernels appear very
structured, with sharp excursions in their values for different pairs
of examples. However, while the good-path kernel structure appears
to be aligned with the task, well distinguishing the two classes,
the adversarial-path kernels structure appears random w.r.t. to the
task. We can expect that adding these adversarial kernels to the good
one would destroy it's task-relevant structure, as can be visually
understood from the total GP kernel. In contrast, the total renormalized
kernel, in which the adversarial-path kernels do not contribute, preserves
the task-relevant structure. Differently, the denoising-path kernel
appears less structured and more uniform, with weaker noisy excursions.
In fact, what we suspect is that there is nothing special about the
specific realization of the random head involved in the denoising
path, which is just implementing a noisy version of uniform attention.
We verify this by substituting the random head with one implementing
uniform attention and repeating the same experiment shown in \Figref{fig:markov}(f) in the main text. This is shown in \Figref{fig:markov_appendix}(c), were we plot
the classification accuracy of the network consisting of the good
and denoising paths alone, for the case of $\sigma_{\parallel}=1$
and $\sigma_{\perp}=5$. We can see that the results are completely
analogous to those shown in \Figref{fig:markov}(f) in the main text.

\section{One-shot image classification task\label{sec:One-shot-appendix}}

\subsection{Training with gradient descent\label{sec:grad-descent-appendix}}
Here we provide details of gradient descent training of our transformer-like model (Sec.~\ref{sec:Model}) for the one-shot image classification task.
We use the Omniglot dataset with the standard 1028/172/432-splits for the train/validation/test class splits \citep{VinyalsBLKW16} as implemented in \texttt{torchmeta} \citep{deleu2019torchmeta}.
We use the Adam optimizer \citep{KingmaB14} using an initial learning rate of $3e^{-4}$ and a batch size of 128 for 10 epochs.
We check the validation accuracy every 1000 steps and select the final model as the one that achieves the best validation accuracy.
We use the binary regression loss as in the theory.
Unlike in the theory, here we train all the model parameters including the key and query projection weight matrices.
We set $N=512$.
All the models considered in this work can be trained on a single A100-40GB GPU within less than 2 hours.

\subsection{Additional results}

\begin{figure}
    \centering    \hspace{-1em}\includegraphics[width=1.0\textwidth]{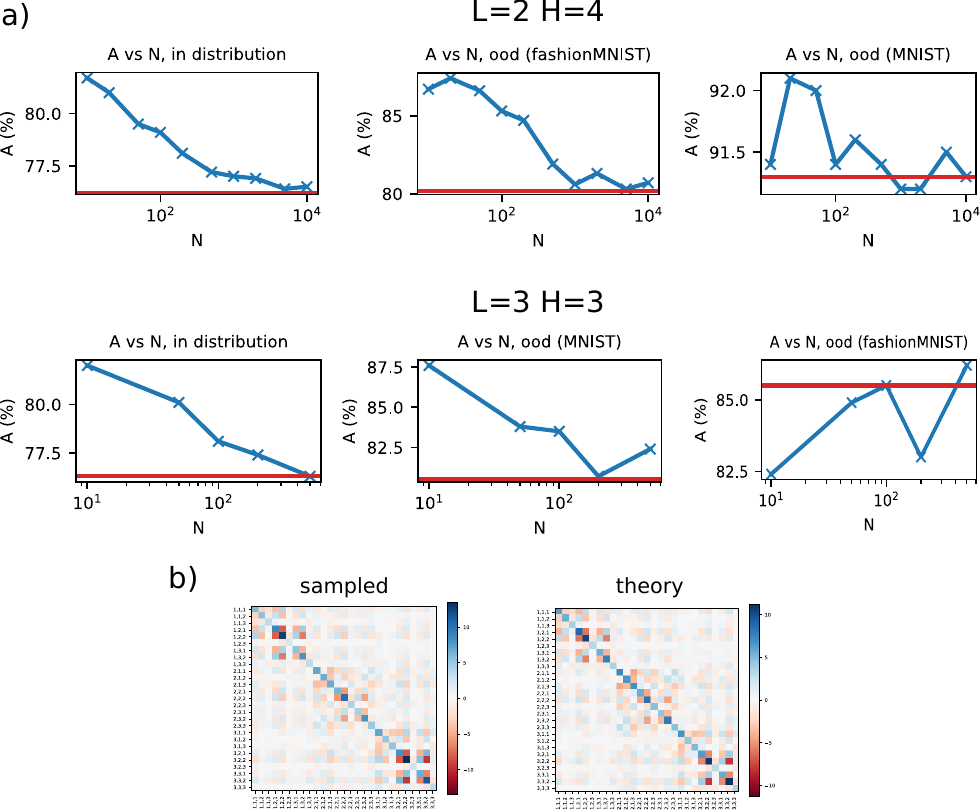}
    \caption{\textbf{(a)} Classification accuracy for varying $N$, tested both in-distribution (Omniglot) and out-of-distribution (MNIST, fashionMNIST). We report the results for the network architecture discussed in the main text ($L=2$, $H=4$) and a deeper one with one more layer ($L=3$, $H=3$). Theory: blue crosses, joined by blue curve. GP limit: red line. \textbf{(b)} Order parameter for the $L=3$, $H=3$ network, for $N=10$. Similarly to the $L=2$, $H=4$ network shown in the main text, the order parameter showcases attention paths interplay, presenting strong off-diagonal elements that deviate from the GP limit.}
    \label{fig:appendix_omniglot_accuracy}
\end{figure}

Here we report further results on the one-shot image classification
task. 

\subsubsection{Optimal temperature\label{apdx_optimal_temperature}}

For the Bayesian model, we find tuning the Gibbs
temperature $\mathcal{T}$ to be particularly important to optimally
perform the task. All results for the network's classification accuracy presented in Sec.~\ref{subsec:One-shot-image-classification} and below
are therefore shown at the optimal temperature for the given $N$,
obtained by scanning the set of temperatures 
$\mathcal{T} \in \{a\,10^{-b}, 1.0, 1.5\}$, where $a\in\left\{1,2.5,5,7.5\right\}$ and $b\in\left\{1,2\right\}$.

Note that temperature is optimized only for the in-distribution classification
accuracy. In particular, we do not optimize for temperature when testing
out-of-distribution, but rather keep the optimal temperature determined
by evaluating the network in-distribution. 

For the network considered in the main text, we find the following optimal temperatures: $\mathcal{T}=0.5$ for $N=10,20,50,100$; $\mathcal{T}=0.25$ for $N=200,500,1000$; $\mathcal{T}=0.1$ for $N=2000,5000$; $\mathcal{T}=0.075$ for $N=10000$.
Note that the optimal
temperature grows consistently as $N$ becomes smaller. This can be
readily understood by inspecting the equation for the mean predictor
(Eq.~\ref{eq:mean_pred}), which we report here for convenience
\[
\left\langle f^{*}\right\rangle =k^{\top}\cdot\left(K+\mathcal{T}\mathbb{I}\right)^{-1}\cdot Y
\]
where we recall that 
\[
K^{\mu\nu}=\frac{1}{H^{L}}\sum_{\pi,\pi'\in\Pi}U^{\pi\pi'}C_{\pi\pi'}^{\mu\nu}\qquad\qquad\mu,\nu=1,\ldots,P
\]
where $C_{\pi\pi'}$ is a path-path kernel. For decreasing $N$, we
typically observe the order parameter growing in overall magnitude,
which in turn affects the magnitude of the kernel $K$. As a consequence,
also the optimal temperature needs to be rescaled. The
fact that $U$ grows in magnitude for smaller $N$ can be understood
from the energy term (Eq.~\ref{eq:apdx_energy_enunciation}) in the action (Eq.~\ref{eq:action}). We discussed
in Sec.~\ref{sec:Theory} that this can be seen as the negative log-likelihood of
the labels vector $Y$ under a centered Gaussian distribution, whose
covariance matrix is the kernel $K$. While the most effective way
to minimize the energy term is that described in the main text (Sec.~\ref{sec:Theory}),
i.e. aligning the kernel with the task, one more trivial way is to
increase the log-likelihood variance in all directions (i.e. increasing
the kernel's overall magnitude). In all of our experiments, we always
observe this phenomenon of growing magnitude to a certain extent.

\subsubsection{Classification accuracy\label{sec:class_accuracy}}

We perform the same experiments on the classification accuracy for
a deeper network ($L=3$, $H=3$). The results are shown in \Figref{fig:appendix_omniglot_accuracy}. Note that
here we also report the test accuracy on MNIST, which was not shown
in the main text. We can see that the results discussed in the main
text are confirmed also for the deeper network. In particular, for
the in-distribution classification accuracy, we consistently observe
a performance improvement in the renormalized regime, with respect
to the GP limit. When testing out of distribution, we can see that
in the best cases (fashionMNIST for ($L=2$, $H=4$); MNIST for ($L=3$,
$H=3$)), the performance improvement is preserved, or at the very
worst (MNIST for ($L=2$, $H=4$); fashionMNIST for ($L=3$, $H=3$))
the GP and renormalized regime show comparable performance.

\subsubsection{Heads pruning\label{sec:appendix_head_pruning}}

\begin{figure}
    \centering    \hspace{-1em}\includegraphics[width=1.0\textwidth]{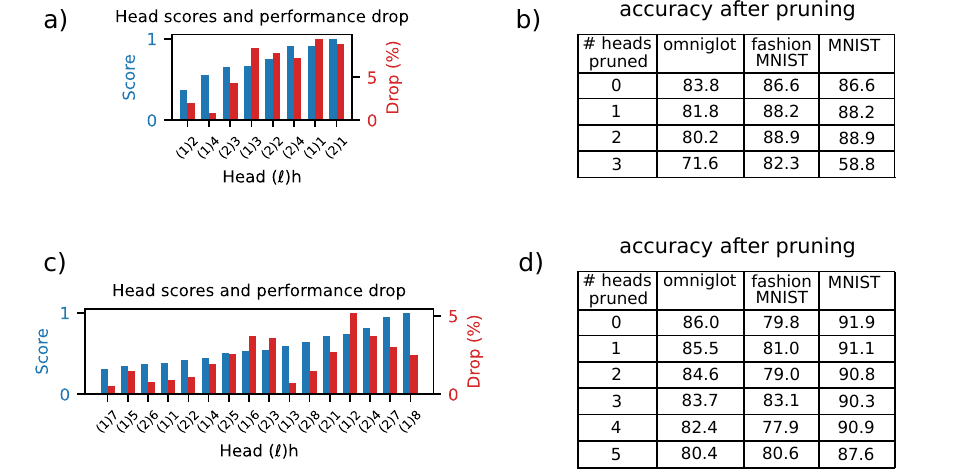}
    \caption{Heads pruning experiment. \textbf{(a,c)} Head score (blue) and performance drop (red) after pruning the head, for the network
trained with gradient descent. (a) smaller network considered in the main text ($L=2$, $H=4$); (c) larger network ($L=2$, $H=8$) \textbf{(b,d)} Classification accuracy of the model trained with gradient descent,
after pruning a growing number of heads, in order of their head score. (b) smaller network considered in the main text ($L=2$, $H=4$); (d) larger network ($L=2$, $H=8$)}
    \label{fig:appendix_omniglot_heads_pruning}
\end{figure}

We repeat the head pruning experiment for a network with more heads
per layer ($L=2$, $H=8$). We find that we can prune with marginal performance loss a similar percentage of heads
as in the smaller network. The results are shown in \Figref{fig:appendix_omniglot_heads_pruning}. Again, we see that the
head scores determined by our theory are qualitatively in line with
the performance loss caused by pruning the corresponding head (\Figref{fig:appendix_omniglot_heads_pruning}(c)).
Note that we do not expect a perfect alignment between the two quantities.
The most important fact to verify is that the low scoring heads correspond
to a small drop in performance. In \Figref{fig:appendix_omniglot_heads_pruning}(d) we show the classification
accuracy after pruning an increasing number of heads, in order of
their score. Up to 4 heads ($25\%$ of the network size) the in-distribution
performance has only a marginal drop, identical to that obtained after pruning
2 heads in the smaller network considered in the main text (also accounting to $25\%$ of its
size, \Figref{fig:appendix_omniglot_heads_pruning}(b)). We can also see that pruning up to three heads improves the
out-of-ditribution performance on fashionMNIST, as we also observed
for the smaller network.


\begin{thebibliography}{68}
\providecommand{\natexlab}[1]{#1}
\providecommand{\url}[1]{\texttt{#1}}
\expandafter\ifx\csname urlstyle\endcsname\relax
  \providecommand{\doi}[1]{doi: #1}\else
  \providecommand{\doi}{doi: \begingroup \urlstyle{rm}\Url}\fi

\bibitem[Vaswani et~al.(2017)Vaswani, Shazeer, Parmar, Uszkoreit, Jones, Gomez, Kaiser, and Polosukhin]{vaswani2017attention}
Ashish Vaswani, Noam Shazeer, Niki Parmar, Jakob Uszkoreit, Llion Jones, Aidan~N Gomez, {\L}ukasz Kaiser, and Illia Polosukhin.
\newblock Attention is all you need.
\newblock In \emph{Proc. Advances in Neural Information Processing Systems (NIPS)}, pages 5998--6008, Long Beach, {CA}, {USA}, December 2017.

\bibitem[Kim et~al.(2017)Kim, Denton, Hoang, and Rush]{kim2016structured}
Yoon Kim, Carl Denton, Luong Hoang, and Alexander~M. Rush.
\newblock Structured attention networks.
\newblock In \emph{Int. Conf. on Learning Representations (ICLR)}, Toulon, France, April 2017.

\bibitem[Parikh et~al.(2016)Parikh, T{\"{a}}ckstr{\"{o}}m, Das, and Uszkoreit]{ParikhT0U16}
Ankur~P. Parikh, Oscar T{\"{a}}ckstr{\"{o}}m, Dipanjan Das, and Jakob Uszkoreit.
\newblock A decomposable attention model for natural language inference.
\newblock In \emph{Proc. Conf. on Empirical Methods in Natural Language Processing (EMNLP)}, pages 2249--2255, Austin, {TX}, {USA}, November 2016.

\bibitem[Cheng et~al.(2016)Cheng, Dong, and Lapata]{cheng16}
Jianpeng Cheng, Li~Dong, and Mirella Lapata.
\newblock Long short-term memory-networks for machine reading.
\newblock In \emph{Proc. Conf. on Empirical Methods in Natural Language Processing (EMNLP)}, pages 551--561, Austin, {TX}, {USA}, November 2016.

\bibitem[Lin et~al.(2017)Lin, Feng, Santos, Yu, Xiang, Zhou, and Bengio]{lin2017structured}
Zhouhan Lin, Minwei Feng, Cicero Nogueira~dos Santos, Mo~Yu, Bing Xiang, Bowen Zhou, and Yoshua Bengio.
\newblock A structured self-attentive sentence embedding.
\newblock In \emph{Int. Conf. on Learning Representations (ICLR)}, Toulon, France, April 2017.

\bibitem[Bahdanau et~al.(2015)Bahdanau, Cho, and Bengio]{bahdanau2014neural}
Dzmitry Bahdanau, Kyunghyun Cho, and Yoshua Bengio.
\newblock Neural machine translation by jointly learning to align and translate.
\newblock In \emph{Int. Conf. on Learning Representations (ICLR)}, San Diego, {CA}, {USA}, May 2015.

\bibitem[Brown et~al.(2020)]{gpt3}
Tom~B Brown et~al.
\newblock Language models are few-shot learners.
\newblock In \emph{Proc. Advances in Neural Information Processing Systems (NeurIPS)}, Virtual only, December 2020.

\bibitem[Dosovitskiy et~al.(2021)Dosovitskiy, Beyer, Kolesnikov, Weissenborn, Zhai, Unterthiner, Dehghani, Minderer, Heigold, Gelly, Uszkoreit, and Houlsby]{dosovitskiy2020image}
Alexey Dosovitskiy, Lucas Beyer, Alexander Kolesnikov, Dirk Weissenborn, Xiaohua Zhai, Thomas Unterthiner, Mostafa Dehghani, Matthias Minderer, Georg Heigold, Sylvain Gelly, Jakob Uszkoreit, and Neil Houlsby.
\newblock An image is worth 16x16 words: Transformers for image recognition at scale.
\newblock In \emph{Int. Conf. on Learning Representations (ICLR)}, Virtual only, May 2021.

\bibitem[Devlin et~al.(2019)Devlin, Chang, Lee, and Toutanova]{devlin2019bert}
Jacob Devlin, Ming{-}Wei Chang, Kenton Lee, and Kristina Toutanova.
\newblock {BERT:} pre-training of deep bidirectional transformers for language understanding.
\newblock In \emph{Proc. North American Chapter of the Association for Computational Linguistics on Human Language Technologies (NAACL-HLT)}, pages 4171--4186, Minneapolis, {MN}, {USA}, June 2019.

\bibitem[Fu et~al.(2023)Fu, Guo, Bai, and Mei]{fu2024can}
Hengyu Fu, Tianyu Guo, Yu~Bai, and Song Mei.
\newblock What can a single attention layer learn? {A} study through the random features lens.
\newblock In \emph{Proc. Advances in Neural Information Processing Systems (NeurIPS)}, New Orleans, LA, USA, December 2023.

\bibitem[Sahiner et~al.(2022)Sahiner, Ergen, Ozturkler, Pauly, Mardani, and Pilanci]{sahiner2022unraveling}
Arda Sahiner, Tolga Ergen, Batu Ozturkler, John~M. Pauly, Morteza Mardani, and Mert Pilanci.
\newblock Unraveling attention via convex duality: Analysis and interpretations of vision transformers.
\newblock In \emph{Proc. Int. Conf. on Machine Learning (ICML)}, Baltimore, Maryland, {USA}, July 2022.

\bibitem[Tarzanagh et~al.(2023)Tarzanagh, Li, Zhang, and Oymak]{tarzanagh2023max}
Davoud~Ataee Tarzanagh, Yingcong Li, Xuechen Zhang, and Samet Oymak.
\newblock Max-margin token selection in attention mechanism.
\newblock In \emph{Proc. Advances in Neural Information Processing Systems (NeurIPS)}, New Orleans, LA, USA, December 2023.

\bibitem[Cui et~al.(2024)Cui, Behrens, Krzakala, and Zdeborov{\'a}]{cui2024phase}
Hugo Cui, Freya Behrens, Florent Krzakala, and Lenka Zdeborov{\'a}.
\newblock A phase transition between positional and semantic learning in a solvable model of dot-product attention.
\newblock \emph{Preprint arXiv:2402.03902}, 2024.

\bibitem[Rende et~al.(2024)Rende, Gerace, Laio, and Goldt]{rende24}
Riccardo Rende, Federica Gerace, Alessandro Laio, and Sebastian Goldt.
\newblock Mapping of attention mechanisms to a generalized potts model.
\newblock \emph{Phys. Rev. Res.}, 6:\penalty0 023057, Apr 2024.

\bibitem[Noci et~al.(2023)Noci, Li, Li, He, Hofmann, Maddison, and Roy]{noci2024shaped}
Lorenzo Noci, Chuning Li, Mufan~Bill Li, Bobby He, Thomas Hofmann, Chris~J. Maddison, and Dan Roy.
\newblock The shaped transformer: Attention models in the infinite depth-and-width limit.
\newblock In \emph{Proc. Advances in Neural Information Processing Systems (NeurIPS)}, New Orleans, LA, USA, December 2023.

\bibitem[Li et~al.(2023{\natexlab{a}})Li, Wang, Liu, and Chen]{li2022theoretical}
Hongkang Li, Meng Wang, Sijia Liu, and Pin{-}Yu Chen.
\newblock A theoretical understanding of shallow vision transformers: Learning, generalization, and sample complexity.
\newblock In \emph{Int. Conf. on Learning Representations (ICLR)}, Kigali, Rwanda, May 2023{\natexlab{a}}.

\bibitem[Boix-Adser{\`a} et~al.(2023)Boix-Adser{\`a}, Littwin, Abbe, Bengio, and Susskind]{boix2023transformers}
Enric Boix-Adser{\`a}, Etai Littwin, Emmanuel Abbe, Samy Bengio, and Joshua~M Susskind.
\newblock Transformers learn through gradual rank increase.
\newblock In Alice Oh, Tristan Naumann, Amir Globerson, Kate Saenko, Moritz Hardt, and Sergey Levine, editors, \emph{Proc. Advances in Neural Information Processing Systems (NeurIPS)}, New Orleans, LA, USA, December 2023.

\bibitem[Tian et~al.(2023)Tian, Wang, Chen, and Du]{tian2023scan}
Yuandong Tian, Yiping Wang, Beidi Chen, and Simon~S. Du.
\newblock Scan and {S}nap: Understanding training dynamics and token composition in 1-layer transformer.
\newblock In \emph{Proc. Advances in Neural Information Processing Systems (NeurIPS)}, New Orleans, LA, USA, December 2023.

\bibitem[Geshkovski et~al.(2023)Geshkovski, Letrouit, Polyanskiy, and Rigollet]{geshkovski2023mathematical}
Borjan Geshkovski, Cyril Letrouit, Yury Polyanskiy, and Philippe Rigollet.
\newblock A mathematical perspective on transformers.
\newblock \emph{Preprint arXiv:2312.10794}, 2023.

\bibitem[Li et~al.(2024)Li, Wang, Lu, Cui, and Chen]{li2024}
Hongkang Li, Meng Wang, Songtao Lu, Xiaodong Cui, and Pin-Yu Chen.
\newblock How do nonlinear transformers learn and generalize in in-context learning?
\newblock In \emph{Proc. Int. Conf. on Machine Learning (ICML)}, Vienna, Austria, July 2024.

\bibitem[Hahn(2020)]{hahn2020theoretical}
Michael Hahn.
\newblock Theoretical limitations of self-attention in neural sequence models.
\newblock \emph{Transactions of the Association for Computational Linguistics}, 8:\penalty0 156--171, 2020.

\bibitem[Edelman et~al.(2022)Edelman, Goel, Kakade, and Zhang]{edelman2022inductive}
Benjamin~L. Edelman, Surbhi Goel, Sham~M. Kakade, and Cyril Zhang.
\newblock Inductive biases and variable creation in self-attention mechanisms.
\newblock In \emph{Proc. Int. Conf. on Machine Learning (ICML)}, Baltimore, MD, {USA}, July 2022.

\bibitem[Nichani et~al.(2024)Nichani, Damian, and Lee]{nichani2024}
Eshaan Nichani, Alex Damian, and Jason~D. Lee.
\newblock How transformers learn causal structure with gradient descent.
\newblock In \emph{Proc. Int. Conf. on Machine Learning (ICML)}, Vienna, Austria, July 2024.

\bibitem[Reddy(2024)]{reddy2024}
Gautam Reddy.
\newblock The mechanistic basis of data dependence and abrupt learning in an in-context classification task.
\newblock In \emph{Int. Conf. on Learning Representations (ICLR)}, Vienna, Austria, May 2024.

\bibitem[Hron et~al.(2020)Hron, Bahri, Sohl{-}Dickstein, and Novak]{hron2020infinite}
Jiri Hron, Yasaman Bahri, Jascha Sohl{-}Dickstein, and Roman Novak.
\newblock Infinite attention: {NNGP} and {NTK} for deep attention networks.
\newblock In \emph{Proc. Int. Conf. on Machine Learning (ICML)}, pages 4376--4386, Virtual only, July 2020.

\bibitem[Lavie et~al.(2024)Lavie, Gur-Ari, and Ringel]{lavie2024towards}
Itay Lavie, Guy Gur-Ari, and Zohar Ringel.
\newblock Towards understanding inductive bias in transformers: A view from infinity.
\newblock \emph{Preprint arXiv:2402.05173}, 2024.

\bibitem[Lee et~al.(2018)Lee, Bahri, Novak, Schoenholz, Pennington, and Sohl{-}Dickstein]{lee2018deep}
Jaehoon Lee, Yasaman Bahri, Roman Novak, Samuel~S. Schoenholz, Jeffrey Pennington, and Jascha Sohl{-}Dickstein.
\newblock Deep neural networks as {G}aussian processes.
\newblock In \emph{Int. Conf. on Learning Representations (ICLR)}, Vancouver, Canada, April 2018.

\bibitem[de~G.~Matthews et~al.(2018)de~G.~Matthews, Hron, Rowland, Turner, and Ghahramani]{matthews2018gaussian}
Alexander~G. de~G.~Matthews, Jiri Hron, Mark Rowland, Richard~E. Turner, and Zoubin Ghahramani.
\newblock Gaussian process behaviour in wide deep neural networks.
\newblock In \emph{Int. Conf. on Learning Representations (ICLR)}, Vancouver, Canada, April 2018.

\bibitem[Li and Sompolinsky(2021)]{li2021statistical}
Qianyi Li and Haim Sompolinsky.
\newblock Statistical mechanics of deep linear neural networks: The backpropagating kernel renormalization.
\newblock \emph{Physical Review X}, 11\penalty0 (3):\penalty0 031059, 2021.

\bibitem[Hanin and Zlokapa(2023)]{hanin2023bayesian}
Boris Hanin and Alexander Zlokapa.
\newblock Bayesian interpolation with deep linear networks.
\newblock \emph{Proc. of the National Academy of Sciences (PNAS)}, 120\penalty0 (23), May 2023.

\bibitem[Pacelli et~al.(2023)Pacelli, Ariosto, Pastore, Ginelli, Gherardi, and Rotondo]{pacelli2023statistical}
R~Pacelli, S~Ariosto, M~Pastore, F~Ginelli, M~Gherardi, and P~Rotondo.
\newblock A statistical mechanics framework for bayesian deep neural networks beyond the infinite-width limit.
\newblock \emph{Nature Machine Intelligence}, 5\penalty0 (12):\penalty0 1497--1507, 2023.

\bibitem[Cui et~al.(2023)Cui, Krzakala, and Zdeborova]{pmlr-v202-cui23b}
Hugo Cui, Florent Krzakala, and Lenka Zdeborova.
\newblock {B}ayes-optimal learning of deep random networks of extensive-width.
\newblock In \emph{Proc. Int. Conf. on Machine Learning (ICML)}, Honolulu, HI, {USA}, July 2023.

\bibitem[Li and Sompolinsky(2022)]{li2022globally}
Qianyi Li and Haim Sompolinsky.
\newblock Globally gated deep linear networks.
\newblock In \emph{Proc. Advances in Neural Information Processing Systems (NeurIPS)}, New Orleans, LA, USA, November 2022.

\bibitem[Belkin et~al.(2019)Belkin, Hsu, Ma, and Mandal]{belkin2019reconciling}
Mikhail Belkin, Daniel Hsu, Siyuan Ma, and Soumik Mandal.
\newblock Reconciling modern machine-learning practice and the classical bias--variance trade-off.
\newblock \emph{Proc. of the National Academy of Sciences (PNAS)}, 116\penalty0 (32):\penalty0 15849--15854, 2019.

\bibitem[Zhang et~al.(2017)Zhang, Bengio, Hardt, Recht, and Vinyals]{zhang2017understanding}
Chiyuan Zhang, Samy Bengio, Moritz Hardt, Benjamin Recht, and Oriol Vinyals.
\newblock Understanding deep learning requires rethinking generalization.
\newblock In \emph{Int. Conf. on Learning Representations (ICLR)}, Toulon, France, April 2017.

\bibitem[Naftali et~al.(1989)Naftali, Esther, and Solla]{tishby1989consistent}
Tishby Naftali, Levin Esther, and Sara~A Solla.
\newblock Consistent inference of probabilities in layered networks: predictions and generalizations.
\newblock In \emph{Proc. Int. Joint Conf. on Neural Networks ({IJCNN})}, pages 403--409, Washington, DC, USA, June 1989.

\bibitem[MacKay(1992)]{mackay1992practical}
David~JC MacKay.
\newblock A practical {B}ayesian framework for backpropagation networks.
\newblock \emph{Neural computation}, 4\penalty0 (3):\penalty0 448--472, 1992.

\bibitem[Neal(1996)]{neal2012bayesian}
Radford~M Neal.
\newblock \emph{Bayesian learning for neural networks}.
\newblock Springer, 1996.

\bibitem[Canatar et~al.(2021)Canatar, Bordelon, and Pehlevan]{canatar2021spectral}
Abdulkadir Canatar, Blake Bordelon, and Cengiz Pehlevan.
\newblock Spectral bias and task-model alignment explain generalization in kernel regression and infinitely wide neural networks.
\newblock \emph{Nature communications}, 12\penalty0 (1):\penalty0 2914, 2021.

\bibitem[Lake et~al.(2015)Lake, Salakhutdinov, and Tenenbaum]{omniglot}
Brenden~M Lake, Ruslan Salakhutdinov, and Joshua~B Tenenbaum.
\newblock Human-level concept learning through probabilistic program induction.
\newblock \emph{Science}, 350\penalty0 (6266):\penalty0 1332--1338, 2015.

\bibitem[LeCun et~al.(1998)LeCun, Cortes, and Burges]{lecun1998mnist}
Yann LeCun, Corinna Cortes, and Christopher~JC Burges.
\newblock The {MNIST} database of handwritten digits.
\newblock URL http://yann.lecun.com/exdb/mnist, 1998.

\bibitem[Xiao et~al.(2017)Xiao, Rasul, and Vollgraf]{xiao2017fashion}
Han Xiao, Kashif Rasul, and Roland Vollgraf.
\newblock Fashion-{MNIST}: a novel image dataset for benchmarking machine learning algorithms.
\newblock \emph{Preprint arXiv:1708.07747}, 2017.

\bibitem[Zhang and Sompolinsky(2024)]{zhang2024robust}
Zechen Zhang and Haim Sompolinsky.
\newblock Robust learning in bayesian parallel branching graph neural networks: The narrow width limit.
\newblock \emph{Preprint arXiv:2407.18807}, 2024.

\bibitem[Voita et~al.(2019)Voita, Talbot, Moiseev, Sennrich, and Titov]{voita2019analyzing}
Elena Voita, David Talbot, Fedor Moiseev, Rico Sennrich, and Ivan Titov.
\newblock Analyzing multi-head self-attention: Specialized heads do the heavy lifting, the rest can be pruned.
\newblock In \emph{Proc. Association for Computational Linguistics (ACL)}, pages 5797--5808, Florence, Italy, July 2019.

\bibitem[Liu et~al.(2023)Liu, Wang, Dao, Zhou, Yuan, Song, Shrivastava, Zhang, Tian, Re, and Chen]{liu23am}
Zichang Liu, Jue Wang, Tri Dao, Tianyi Zhou, Binhang Yuan, Zhao Song, Anshumali Shrivastava, Ce~Zhang, Yuandong Tian, Christopher Re, and Beidi Chen.
\newblock Deja vu: Contextual sparsity for efficient {LLM}s at inference time.
\newblock In \emph{Proc. Int. Conf. on Machine Learning (ICML)}, Honolulu, Hawaii, {USA}, July 2023.

\bibitem[Touvron et~al.(2023)Touvron, Lavril, Izacard, Martinet, Lachaux, Lacroix, Rozi{\`{e}}re, Goyal, Hambro, Azhar, Rodriguez, Joulin, Grave, and Lample]{touvron2023llama}
Hugo Touvron, Thibaut Lavril, Gautier Izacard, Xavier Martinet, Marie{-}Anne Lachaux, Timoth{\'{e}}e Lacroix, Baptiste Rozi{\`{e}}re, Naman Goyal, Eric Hambro, Faisal Azhar, Aur{\'{e}}lien Rodriguez, Armand Joulin, Edouard Grave, and Guillaume Lample.
\newblock Llama: Open and efficient foundation language models.
\newblock \emph{Preprint arXiv:2302.13971}, 2023.

\bibitem[Eldan and Li(2023)]{eldan2023tinystories}
Ronen Eldan and Yuanzhi Li.
\newblock Tinystories: How small can language models be and still speak coherent english?
\newblock \emph{Preprint arXiv:2305.07759}, 2023.

\bibitem[Geiger et~al.(2020)Geiger, Spigler, Jacot, and Wyart]{geiger2020disentangling}
Mario Geiger, Stefano Spigler, Arthur Jacot, and Matthieu Wyart.
\newblock Disentangling feature and lazy training in deep neural networks.
\newblock \emph{Journal of Statistical Mechanics: Theory and Experiment}, 2020\penalty0 (11):\penalty0 113301, 2020.

\bibitem[Bordelon and Pehlevan(2023{\natexlab{a}})]{NEURIPS2023_1ec69275}
Blake Bordelon and Cengiz Pehlevan.
\newblock Dynamics of finite width kernel and prediction fluctuations in mean field neural networks.
\newblock In \emph{Proc. Advances in Neural Information Processing Systems (NeurIPS)}, New Orleans, LA, USA, December 2023{\natexlab{a}}.

\bibitem[van Meegen and Sompolinsky(2024)]{vanmeegen2024codingschemesneuralnetworks}
Alexander van Meegen and Haim Sompolinsky.
\newblock Coding schemes in neural networks learning classification tasks.
\newblock \emph{Preprint arXiv:2407.18807}, 2024.

\bibitem[Saxe et~al.(2019)Saxe, McClelland, and Ganguli]{saxe2019mathematical}
Andrew~M Saxe, James~L McClelland, and Surya Ganguli.
\newblock A mathematical theory of semantic development in deep neural networks.
\newblock \emph{Proc. of the National Academy of Sciences (PNAS)}, 116\penalty0 (23):\penalty0 11537--11546, 2019.

\bibitem[Bordelon and Pehlevan(2023{\natexlab{b}})]{bordelon2023self}
Blake Bordelon and Cengiz Pehlevan.
\newblock Self-consistent dynamical field theory of kernel evolution in wide neural networks.
\newblock \emph{Journal of Statistical Mechanics: Theory and Experiment}, 2023\penalty0 (11):\penalty0 114009, 2023{\natexlab{b}}.

\bibitem[Avidan et~al.(2023)Avidan, Li, and Sompolinsky]{avidan2023connecting}
Yehonatan Avidan, Qianyi Li, and Haim Sompolinsky.
\newblock Connecting {NTK} and {NNGP}: A unified theoretical framework for neural network learning dynamics in the kernel regime.
\newblock \emph{Preprint arXiv:2309.04522}, 2023.

\bibitem[Yun et~al.(2020)Yun, Bhojanapalli, Rawat, Reddi, and Kumar]{yun2019transformers}
Chulhee Yun, Srinadh Bhojanapalli, Ankit~Singh Rawat, Sashank~J. Reddi, and Sanjiv Kumar.
\newblock Are transformers universal approximators of sequence-to-sequence functions?
\newblock In \emph{Int. Conf. on Learning Representations (ICLR)}, Virtual only, April 2020.

\bibitem[Jelassi et~al.(2022)Jelassi, Sander, and Li]{jelassi2022vision}
Samy Jelassi, Michael~E. Sander, and Yuanzhi Li.
\newblock Vision transformers provably learn spatial structure.
\newblock In \emph{Proc. Advances in Neural Information Processing Systems (NeurIPS)}, New Orleans, LA, USA, November 2022.

\bibitem[Bai et~al.(2023)Bai, Chen, Wang, Xiong, and Mei]{bai2024transformers}
Yu~Bai, Fan Chen, Huan Wang, Caiming Xiong, and Song Mei.
\newblock Transformers as statisticians: Provable in-context learning with in-context algorithm selection.
\newblock In \emph{Proc. Advances in Neural Information Processing Systems (NeurIPS)}, New Orleans, LA, USA, December 2023.

\bibitem[Guo et~al.(2024)Guo, Hu, Mei, Wang, Xiong, Savarese, and Bai]{guo2023transformers}
Tianyu Guo, Wei Hu, Song Mei, Huan Wang, Caiming Xiong, Silvio Savarese, and Yu~Bai.
\newblock How do transformers learn in-context beyond simple functions? a case study on learning with representations.
\newblock In \emph{Int. Conf. on Learning Representations (ICLR)}, Vienna, Austria, 2024.

\bibitem[Li et~al.(2023{\natexlab{b}})Li, Ildiz, Papailiopoulos, and Oymak]{li2023transformers}
Yingcong Li, Muhammed~Emrullah Ildiz, Dimitris Papailiopoulos, and Samet Oymak.
\newblock Transformers as algorithms: Generalization and stability in in-context learning.
\newblock In \emph{Proc. Int. Conf. on Machine Learning (ICML)}, Honolulu, HI, {USA}, July 2023{\natexlab{b}}.

\bibitem[Zhang et~al.(2024)Zhang, Frei, and Bartlett]{zhang2023trained}
Ruiqi Zhang, Spencer Frei, and Peter~L Bartlett.
\newblock Trained transformers learn linear models in-context.
\newblock \emph{Journal of Machine Learning Research (JMLR)}, 25\penalty0 (49):\penalty0 1--55, 2024.

\bibitem[Zhang et~al.(2023)Zhang, Zhang, Yang, and Wang]{zhang2023and}
Yufeng Zhang, Fengzhuo Zhang, Zhuoran Yang, and Zhaoran Wang.
\newblock What and how does in-context learning learn? {B}ayesian model averaging, parameterization, and generalization.
\newblock \emph{Preprint arXiv:2305.19420}, 2023.

\bibitem[Fedoryuk(1977)]{fedoryuk1977saddle}
Mikhail~Vasil'evich Fedoryuk.
\newblock \emph{The saddle-point method}.
\newblock Nauka, Moscow, 1977.

\bibitem[Fedoryuk(1989)]{fedoryuk1989asymptotic}
Mikhail~Vasil'evich Fedoryuk.
\newblock Asymptotic methods in analysis.
\newblock In \emph{Analysis I: integral representations and asymptotic methods}, pages 83--191. Springer, 1989.

\bibitem[Kingma and Ba(2015)]{KingmaB14}
Diederik~P. Kingma and Jimmy Ba.
\newblock Adam: {A} method for stochastic optimization.
\newblock In \emph{Int. Conf. on Learning Representations (ICLR)}, San Diego, CA, USA, May 2015.

\bibitem[Betancourt(2018)]{betancourt2018conceptual}
Michael Betancourt.
\newblock A conceptual introduction to {H}amiltonian {M}onte {C}arlo.
\newblock \emph{Preprint arXiv:1701.02434}, 2018.

\bibitem[Phan et~al.(2019)Phan, Pradhan, and Jankowiak]{phan2019composable}
Du~Phan, Neeraj Pradhan, and Martin Jankowiak.
\newblock Composable effects for flexible and accelerated probabilistic programming in {N}um{P}yro.
\newblock \emph{Preprint arXiv:1912.11554}, 2019.

\bibitem[Bingham et~al.(2019)Bingham, Chen, Jankowiak, Obermeyer, Pradhan, Karaletsos, Singh, Szerlip, Horsfall, and Goodman]{bingham2019pyro}
Eli Bingham, Jonathan~P. Chen, Martin Jankowiak, Fritz Obermeyer, Neeraj Pradhan, Theofanis Karaletsos, Rohit Singh, Paul~A. Szerlip, Paul Horsfall, and Noah~D. Goodman.
\newblock Pyro: Deep universal probabilistic programming.
\newblock \emph{Journal of Machine Learning Research (JMLR)}, 20\penalty0 (28):\penalty0 1--6, 2019.

\bibitem[Vinyals et~al.(2016)Vinyals, Blundell, Lillicrap, Kavukcuoglu, and Wierstra]{VinyalsBLKW16}
Oriol Vinyals, Charles Blundell, Tim Lillicrap, Koray Kavukcuoglu, and Daan Wierstra.
\newblock Matching networks for one shot learning.
\newblock In \emph{Proc. Advances in Neural Information Processing Systems (NIPS)}, pages 3630--3638, Barcelona, Spain, December 2016.

\bibitem[Deleu et~al.(2019)Deleu, W{\"u}rfl, Samiei, Cohen, and Bengio]{deleu2019torchmeta}
Tristan Deleu, Tobias W{\"u}rfl, Mandana Samiei, Joseph~Paul Cohen, and Yoshua Bengio.
\newblock Torchmeta: A meta-learning library for {P}y{T}orch.
\newblock \emph{Preprint arXiv:1909.06576}, 2019.

\end{thebibliography}
\end{document}